\pdfoutput=1

\documentclass[11pt]{article}

\usepackage[]{EMNLP2022}

\usepackage{times}
\usepackage{latexsym}

\usepackage[T1]{fontenc}

\usepackage[utf8]{inputenc}

\usepackage{microtype}

\usepackage{inconsolata}

\title{Gender Biases Unexpectedly Fluctuate in the Pre-training Stage of Masked Language Models}

\author{Kenan Tang$^*$ \quad Hanchun Jiang$^*$\\
  University of Chicago \\
  \texttt{\{kenantang, carolinejiang\}@uchicago.edu}\\}

\usepackage{amssymb}

\usepackage{graphicx}

\begin{document}
\maketitle

\begingroup\def\thefootnote{*}\begin{NoHyper}\footnotetext{Equal contribution.}\end{NoHyper}\endgroup

\begin{abstract}

Masked language models pick up gender biases during pre-training. Such biases are usually attributed to a certain model architecture and its pre-training corpora, with the implicit assumption that other variations in the pre-training process, such as the choices of the random seed or the stopping point, have no effect on the biases measured. However, we show that severe fluctuations exist at the fundamental level of individual templates, invalidating the assumption. Further against the intuition of how humans acquire biases, these fluctuations are not correlated with the certainty of the predicted pronouns or the profession frequencies in pre-training corpora. We release our code and data to benefit future research\footnote{\url{https://github.com/kt2k01/checkpoint-bias}}.

\end{abstract}

\section{Introduction}

Masked language models (MLMs) succeed in solving natural language processing tasks, under the paradigm of fine-tuning the publicly released pre-trained checkpoints \cite{devlin-etal-2019-bert, liu2019roberta}. The pre-training process uses large-scale human corpora, a practice that raises the concern whether harmful gender biases in human language are picked up by MLMs. Thus, much effort has been devoted to the quantification of gender biases in these models \cite{delobelle-etal-2022-measuring}.

Meanwhile, the high cost of pre-training usually prohibits researchers from reproducing an MLM from scratch. Therefore, the single public checkpoint of an MLM architecture, such as \texttt{bert-base-uncased}, has become the default choice for bias quantification. The biases are reported as a property of the model identified by its architecture \cite{alnegheimish-etal-2022-using}.

\begin{figure}
    \centering
    \includegraphics[width=7cm]{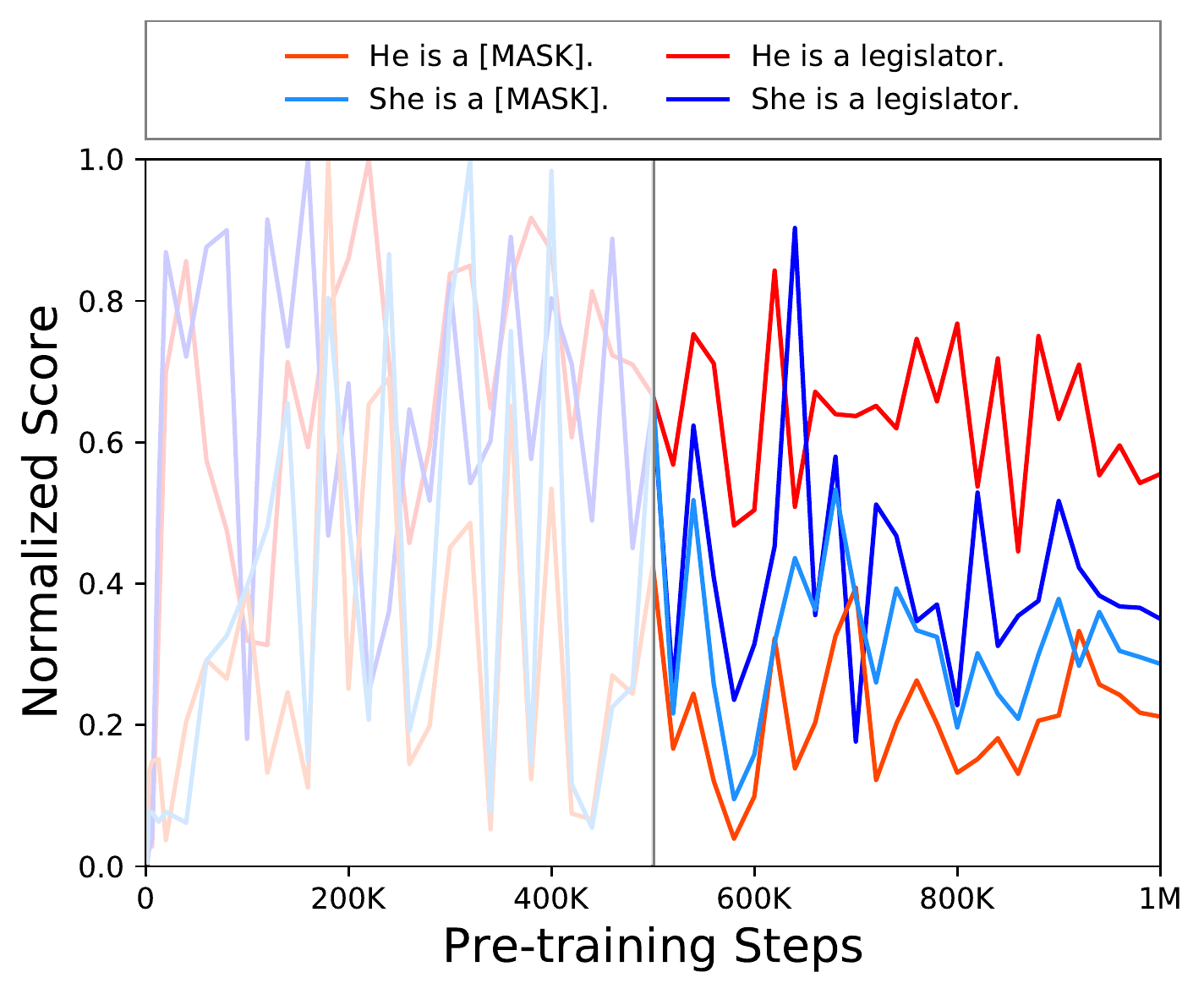}
    \caption{For RoBERTa, The probabilities of filling in gender pronouns into templates change unexpectedly after the training and validation loss has plateaued after 500K steps. Here results from the template for prior estimation and the one with the first profession in the full list are shown. For better visualization, the probability scores have been normalized by the maximal scores during pre-training.}
    \label{fig:intro-figure}
\end{figure}

Inevitably, this reporting scheme understates the potential importance of hyperparameter choices. In this paper, we discuss two hyperparameters, namely the random seed and the number of training steps. Intuitively, they impact pre-training minimally. Changing random seeds should cause a small variance in the model performance, and further pre-training after the loss has plateaued should not change downstream performance a lot. 

However, since the pre-training objective is not bias-aware, it is risky to assume that MLMs are equally biased when the random seed or the number of pre-training steps changes. The risk is amplified by the imperfection of popular gender bias metrics, which involve the probabilities of filling gender pronouns into only a few templates. This risk undermines the validity of applying continued pre-training to identify gender biases in a certain corpus \cite{bertsch-etal-2022-evaluating}.

In this paper, we demonstrate how gender biases fluctuate (Figure \ref{fig:intro-figure}) when the seemingly irrelevant hyperparameters change. Inspired by intuitions about how humans understand gender stereotypes, we then provide fine-grained analyses on templates constructed from individual professions.

The main contributions of this paper are:

\begin{itemize}
    \item By jointly analyzing pre-training dynamics and gender biases, we argue against framing gender biases as an innate property of a model architecture and its pre-training corpora.
    \item By differentiating the behaviors of a model on individual templates without aggregation, we describe in detail of how MLMs behave on this type of gender bias probe.
\end{itemize}

\section{Related Work}

\subsection{Pre-training Dynamics}

Intermediate pre-training checkpoints are needed for investigating the effects of different random seeds and stopping points. Despite the apparent availability, architecture designers seldom release these checkpoints. With limited budget, we can still fortunately rely on a limited number of replication studies where such checkpoints are released for models including BERT \cite{DBLP:conf/iclr/SellamYTWSDLBTE22}, RoBERTa \cite{liu-etal-2021-probing-across}, GPT-2 \cite{Mistral}, and ALBERT \cite{chiang-etal-2020-pretrained}. While these works also probe the pre-training process from different angles, we add the missing discussion on gender bias probes, specifically for the two MLMs BERT (\texttt{bert-base-uncased}) and RoBERTa (\texttt{roberta-base}). 

However, we omit the comparison between the replicated checkpoints and original ones from Hugging Face \cite{wolf2019huggingface}, because the referred papers state that they have failed on exact replications.

\subsection{Gender Bias Metrics}

In static word embedding models, gender biases can be computed from distances between gendered words and non-gendered profession names \cite{staticbias}. In contextual word embedding models, a similar method uses templates constructed from profession names as the input. For both MLMs \cite{delobelle-etal-2022-measuring} and auto-regressive models such as GPT-2 \cite{alnegheimish-etal-2022-using}, output probabilities of generating gender pronouns are divided to get a ratio as the gender bias score. While this is the only approach we use, we refer interested readers to comprehensive surveys in this area (\citealp{stanczak2021survey}).

\section{Methods}

\subsection{The Template-based Approach}

In the template-based approach for gender bias measurement, a model checkpoint at $m$ pre-training steps generate two sentences with different probabilities from one template. An example of such a generated sentence is
$$\texttt{He is a president.}$$
A template $t$ has the four-component form:
$$t=\texttt{[MASK] <VERB> <DET> <PROFESSION>.}$$
The template is constructed by choosing the value of the latter three components. The second component \texttt{<VERB>} is chosen from ``is/works as''. The value of the third component \texttt{<DET>} chosen from ``a/an'' is determined by the initial phoneme of the value of the fourth component \texttt{<PROFESSION>}, chosen from a set constructed by merging lists from \citet{delobelle-etal-2022-measuring} and \citet{alnegheimish-etal-2022-using}. The first list contains 30 male-stereotypical professions and 30 female-sterotypical ones. The second list contains 893 professions scraped from Wikipedia. The merging enables both a comprehensive comparison of many professions and a finer-grained study of gender stereotypes, though the latter is not our focus. The size of the set is 923 after filtering repeated professions.

Taking $t$ as input, the checkpoint $m$ fills in the first component \texttt{[MASK]} with either ``he'' or ``she''. Denote the probabilities as $\mathrm{P}(he|m,t)$ and $\mathrm{P}(she|m,t)$, we define the bias score as
$$r(m,t) = \frac{\mathrm{P}(he|m,t)}{\mathrm{P}(she|m,t)}.$$
A ratio of $r(m,t)=1$ implies that the checkpoint $m$ is fair for the profession in $t$. 

We use a matrix $\mathbf{R} \in \mathbb{R}^{s\times p}$ to denote the collection of $\mathbf{R}_{m,t} = r(m,t)$, where $b=62$ is the number of sampled training steps for RoBERTa ($b=29$ for BERT) and $p=923$ is the number of professions. Without ambiguity, $t$ denotes either a template or the index of the profession in this template. The model choice and random seed will not be indexed in the notation $\mathbf{R}$ but will be specified in context. All indices start from 0.

Upon observing that models output $r(m,t)>1$ for most templates, some previous works have suggested normalizing $r(m,t)$ by the priors of gender pronouns \cite{tal-etal-2022-fewer, alnegheimish-etal-2022-using}. To estimate the priors, we use a template $t_p$ where \texttt{<PROFESSION>} is replaced by a \texttt{[MASK]}, but the checkpoint still only predicts the first \texttt{[MASK]}. Then, the bias score normalized by priors is
$$n(m,t) = r(m,t)\cdot\frac{\mathrm{P}(she|m,t_p)}{\mathrm{P}(he|m,t_p)}.$$
A ratio of $n(m,t)=1$ similarly implies fairness, and we use a matrix $\mathbf{N} \in \mathbb{R}^{s\times p}$ to denote the collection of $n(m,t)$. We will use both definitions.

To quantify whether a template is natural, we define the certainty $c(m,t)$ for a checkpoint $m$ and a template $t$ as $c(m,t) = \mathrm{P}(he|m,t) + \mathrm{P}(she|m,t)$. Higher certainty suggests higher naturalness. A matrix $\mathbf{C} \in \mathbb{R}^{s\times p}$ denotes the collection of  $c(m,t)$.

\subsection{Fluctuations}

On one hand, we would like to quantify fluctuations of $n(m,t)$ when it is not expected to change much at later stages of pre-training. We calculate the coefficient of variation (CV) from incomplete columns $\mathbf{N}_{k:b,t}$ in the matrix $\mathbf{N}$. Each incomplete column starts at the row indexed by $k$, a point when (1) the training and validation loss has already plateaued or (2) the downstream performances after fine-tuning do not improve much with further pre-training. 

For the RoBERTa checkpoints, the plateau is reported to start at 50K steps out of 1M total steps. The number is 1M out of 2M for BERT. We more conservatively take the point of 500K steps for RoBERTa. In other words, we set $k=36$ for RoBERTa and $k=18$ for BERT. The choice is further discussed in Appendix \ref{sec:plateau}. 

The CV is denoted by a vector $\mathbf{v}$, with entries
$$\mathbf{v}_t = \mathrm{CV}(\mathbf{N}_{k:b,t}) = \frac{\mathrm{SD}(\mathbf{N}_{k:b,t})}{\mathrm{AM}(\mathbf{N}_{k:b,t})},$$
where SD is the standard deviation, and AM is the arithmetic mean. We compute the Pearson correlation coefficient between the CV $\mathbf{v}$ and the mean certainties $\mathbf{c}$, with entries
$$\mathbf{c}_t = \mathrm{AM}(\mathbf{C}_{k:b, t}).$$

On the other hand, to study the effect of changing random seeds, we base our investigation on 5 pre-training runs of BERT, as there is only a single run of RoBERTa. For each pair of random seeds, we compute the Pearson correlation coefficient between the pair of the averaged ratios $\mathbf{n}$, with entries
$$\mathbf{n}_t = \mathrm{AM}(\mathbf{N}_{k:b, t}).$$

Additionally, we repeat the above procedure on the unnormalized $\mathbf{R}$ instead of $\mathbf{N}$. Using $\mathbf{R}$, we similarly define vectors $\mathbf{v}$ and $\mathbf{r}$.

\subsection{Frequency in the Corpus}

To identify a potential cause of the fluctuation, we look for the frequency of each profession in the pre-training corpora. Because we cannot count directly, we use frequencies in the BookCorpus as an estimation, since it is a large pre-training corpus shared by BERT and RoBERTa. We first use the Google Ngram API to query the yearly relative frequencies of each profession (case-sensitivity consistent with that of the model). Then, to get the total frequency, we multiply the yearly relative frequencies by the yearly total sizes of the corpus (from 1700 to 2000) released by the curators \cite{bookcorpus}. The corpus is constantly evolving so that we cannot exactly compute the frequencies in the pre-training corpus used by the replication studies, but we believe this estimation is good enough (further discussion in Appendix \ref{sec:freqs}).

The frequencies for all professions are denoted by a vector $\mathbf{f}$, whose entries are the inner product
$$\mathbf{f}_t = \mathbf{s} \cdot \mathbf{y}(t),$$
where the vector $\mathbf{s}$ denotes yearly sizes and $\mathbf{y}(t)$ denotes yearly frequencies of a profession $t$. We compute the Pearson correlation coefficient between the CV $\mathbf{v}$ and the frequencies $\mathbf{f}$.

\begin{figure*}
    \centering
    \includegraphics[width=5cm]{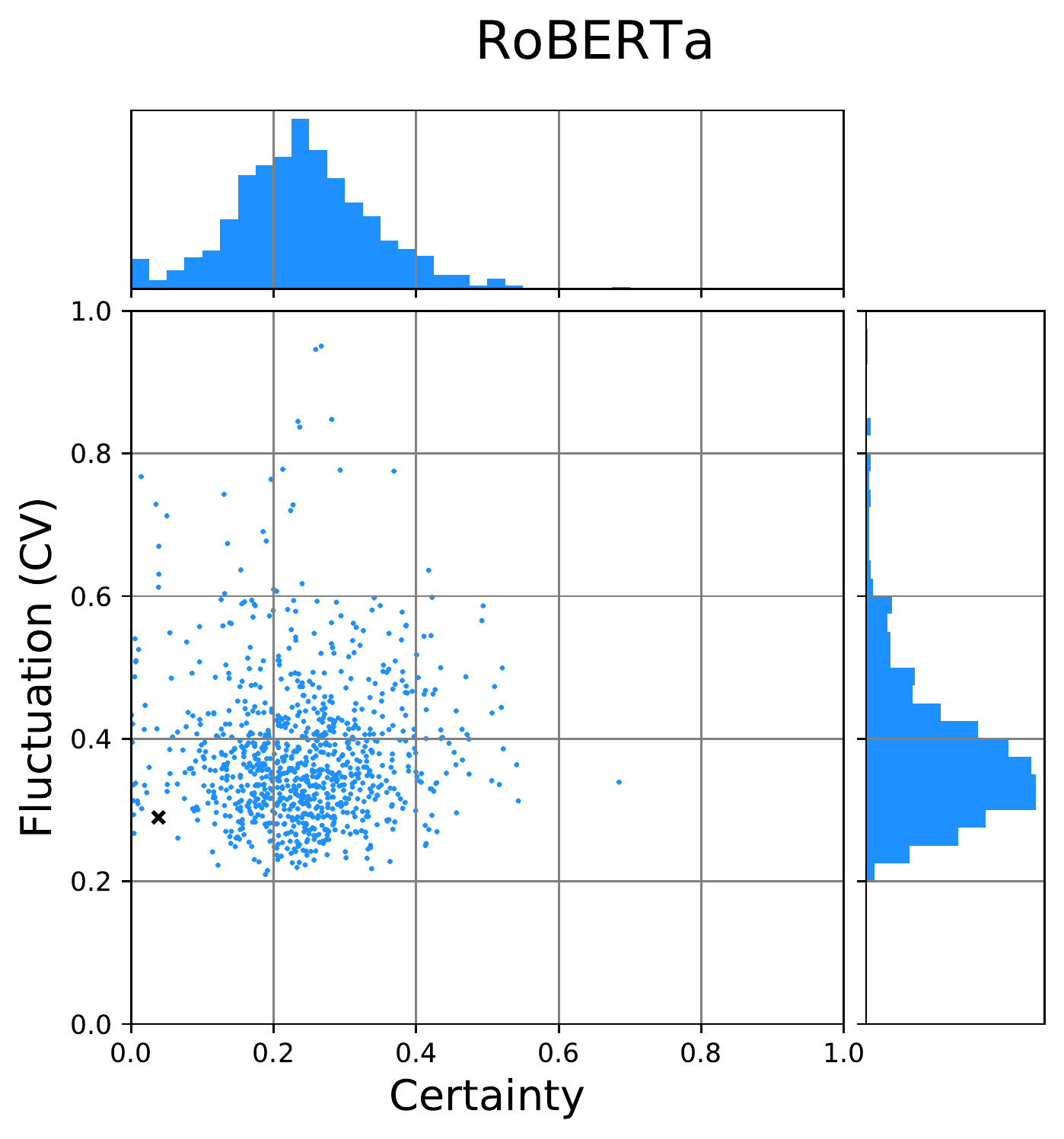}
    \includegraphics[width=5cm]{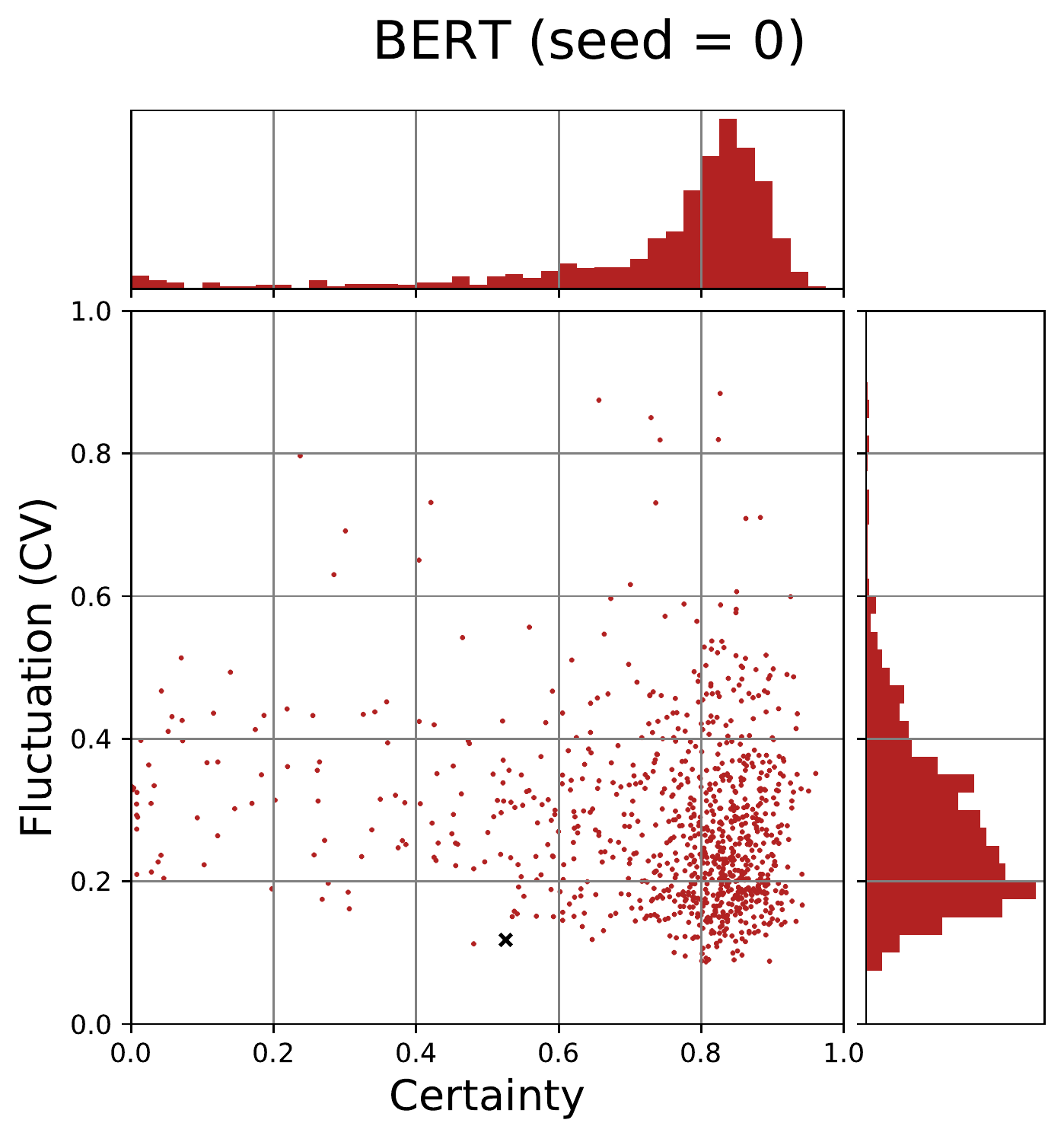}
    \includegraphics[width=5cm]{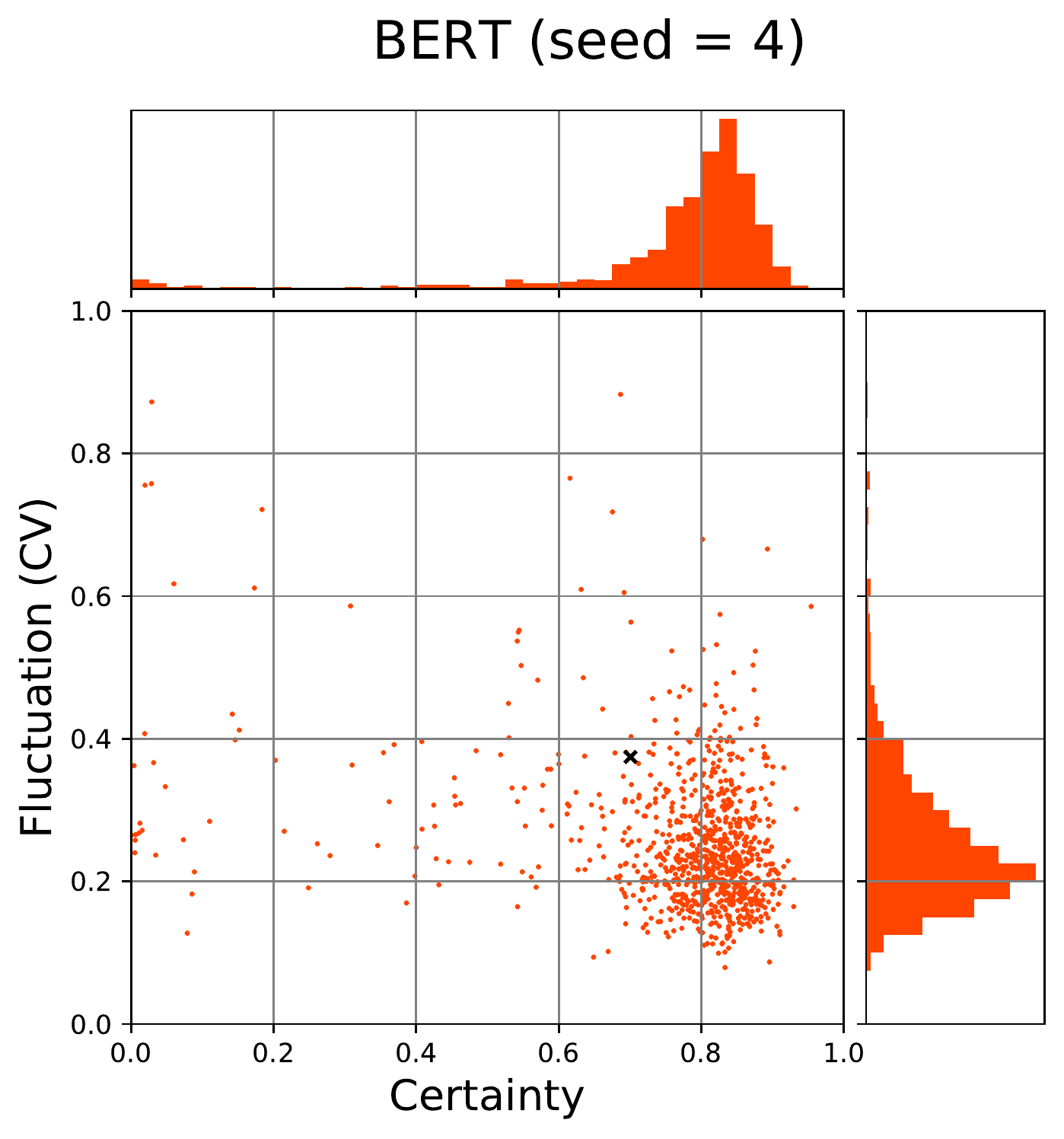}
    \\
    \includegraphics[width=5cm]{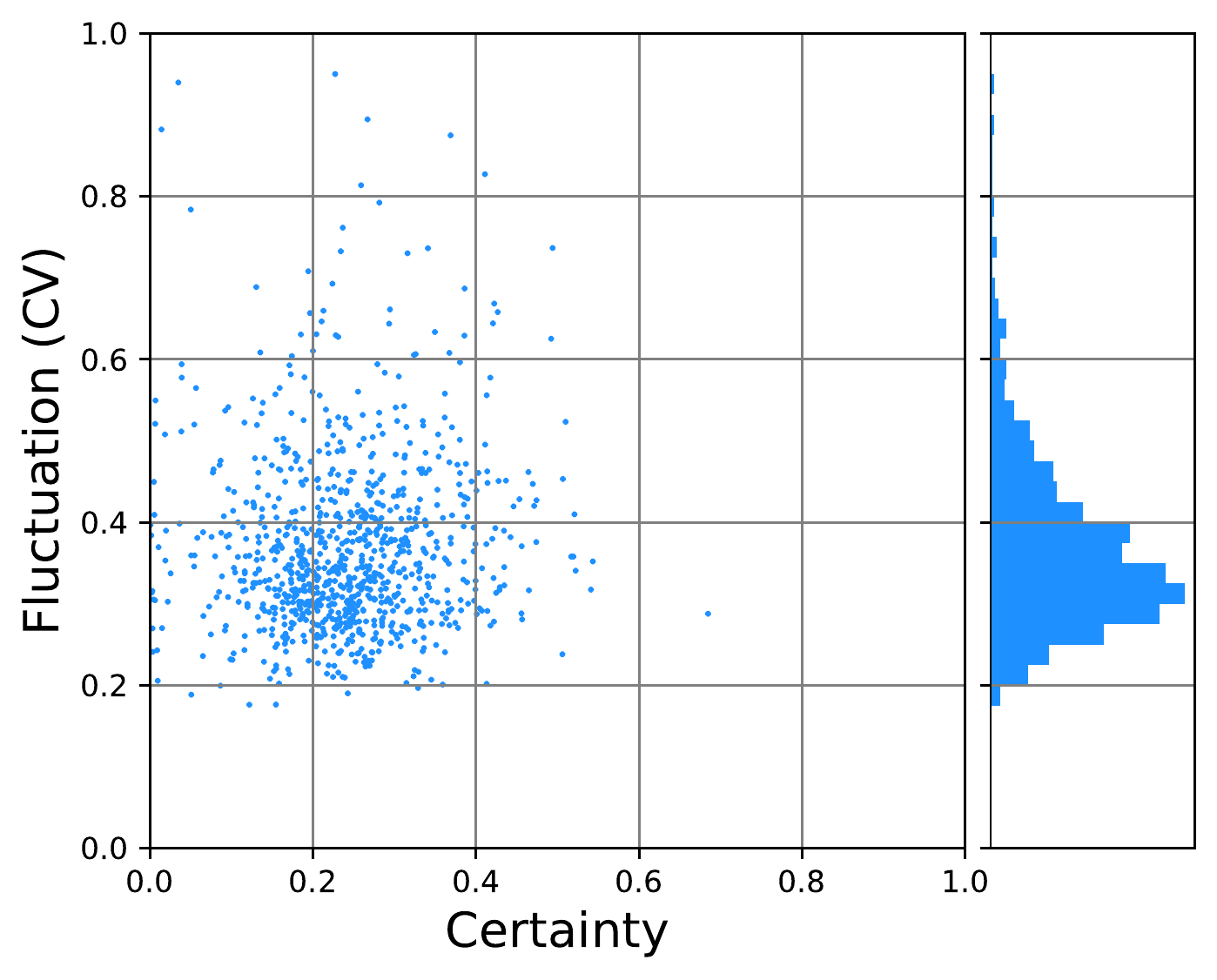}
    \includegraphics[width=5cm]{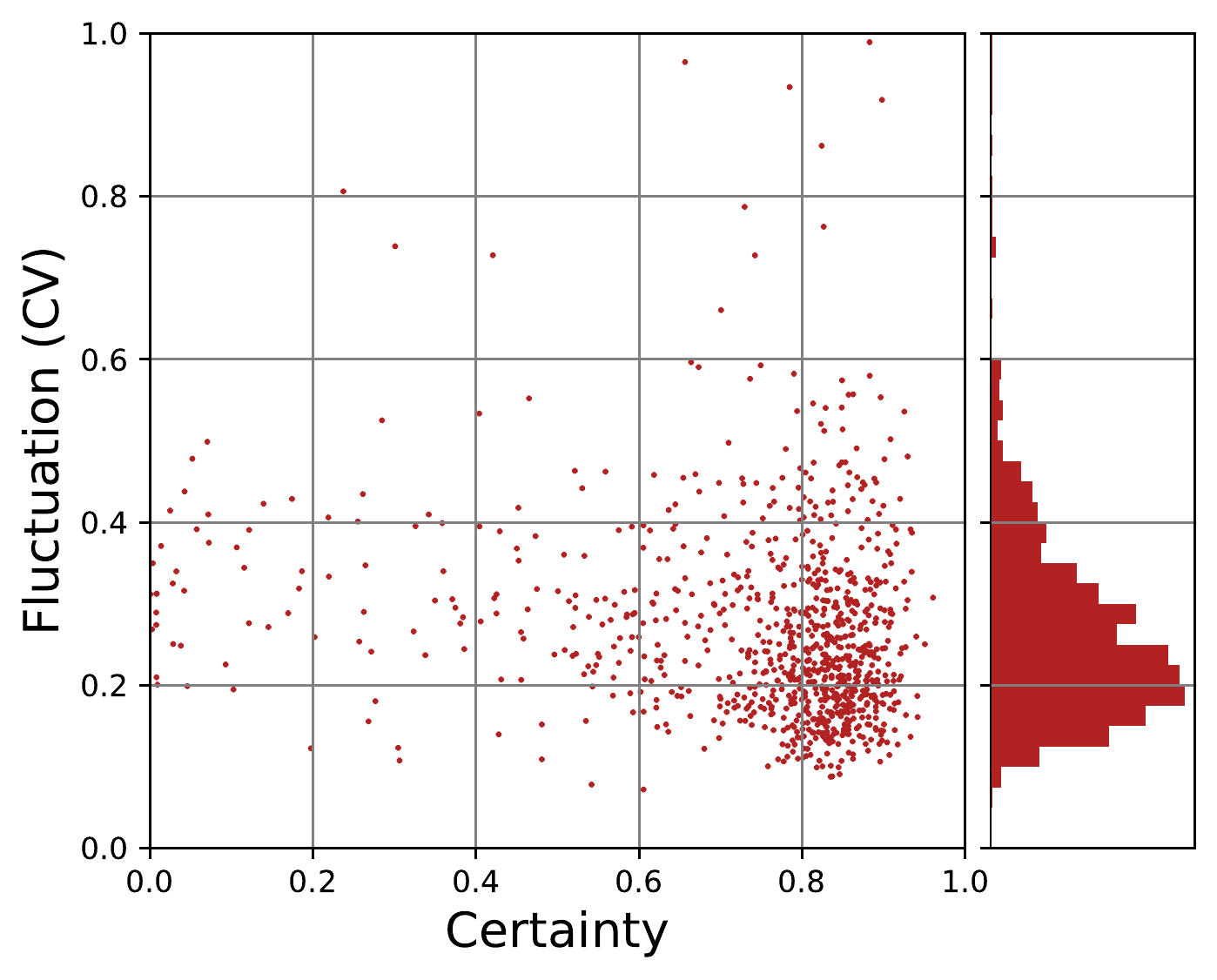}
    \includegraphics[width=5cm]{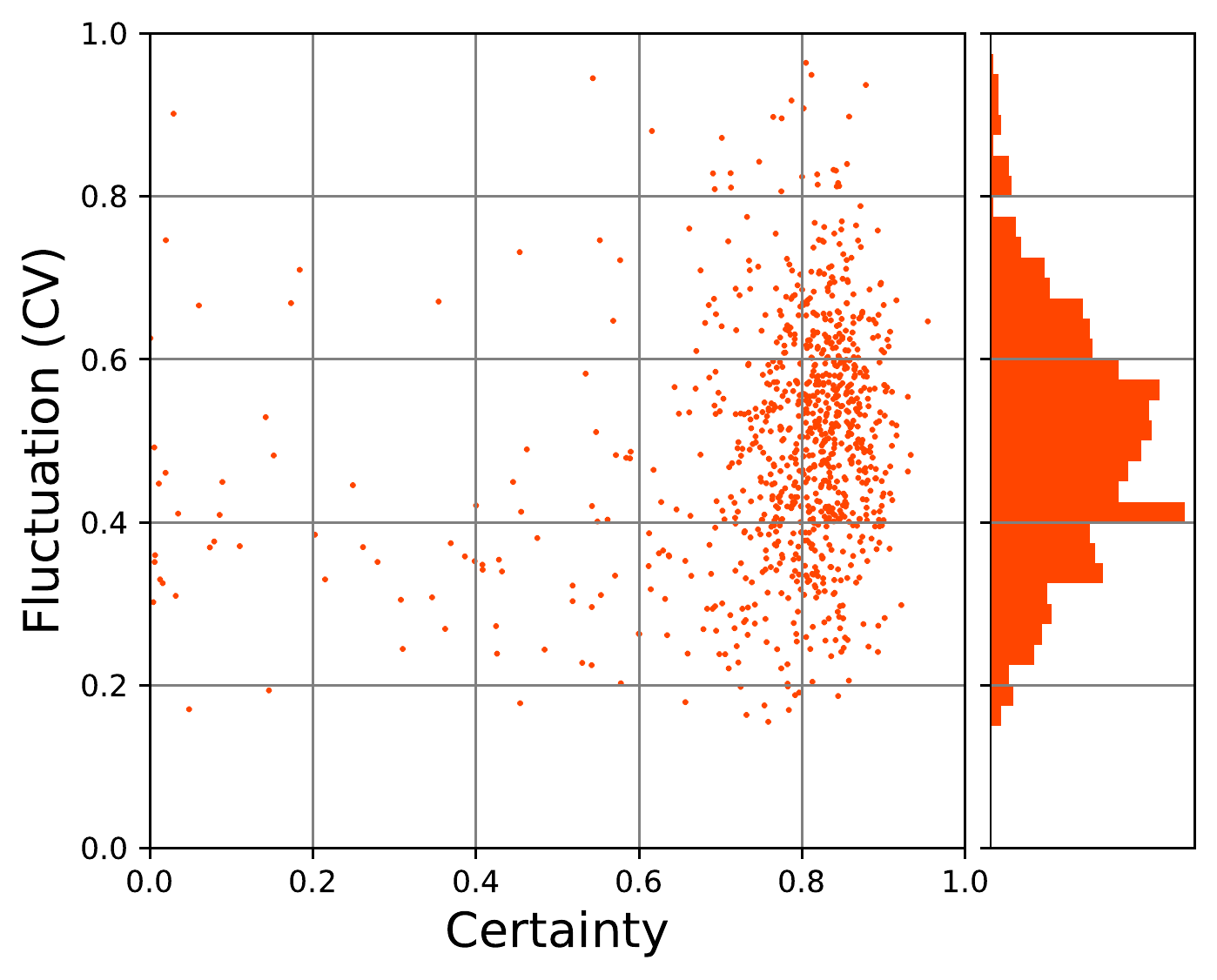}
    \caption{Certainty negligibly correlates with either unnormalized ratios (top row) or normalized ratios (bottom row), for both RoBERTa (first column) and BERT with different random seeds (second and third column). The certainty and fluctuation of the prior is represented by $\times$ on the top 3 plots. Distributions are visualized by histograms.}
    \label{fig:RQ12}
\end{figure*}

\section{Results and Discussion}

We present our results around the following research questions, inspired by intuitions about how models might pick up biases and how this can be analogized to the patterns of how humans hold gender stereotypes \cite{ellemers2018gender}. 

Here, we only base our discussion on the results when the \texttt{<VERB>} in the template is ``is''. The results from the alternative ``works as'' template are qualitatively the same (Appendix \ref{sec:works-as}).

\noindent\textbf{RQ1: Do biases fluctuate, even after the training and validation loss has plateaued?}

Humans' perception of gender stereotypes is relatively fixed. However, for both models, biases for all professions fluctuate considerably after the loss has plateaued (Figure \ref{fig:RQ12}). Take RoBERTa as an example, the smallest fluctuations of unnormalized ratios are still above a certain threshold $\min(\mathbf{v}) = 0.21$, while the highest fluctuations can reach $\max(\mathbf{v}) = 0.95$.

Our result is consistent with the previous finding that model checkpoints in the pre-training process fail to behave consistently under probes of factual knowledge in the form of templates \cite{liu-etal-2021-probing-across}.

\noindent\textbf{RQ2: Do biases fluctuate less if the model is more certain about its predictions?}


With a strong belief in gender stereotypes, a human tends to perform according to the stereotype, reinforcing the stereotype in the process. While the models have distributions of certainties centered at different means, the correlations between the CV $\mathbf{v}$ and the certainties $\mathbf{c}$ are both weak (Figure \ref{fig:RQ12}).

\noindent\textbf{RQ3: Do biases fluctuate less if the model sees the profession less often?}


For humans, gender stereotypes are constantly reproduced as a result of mixed motivations upon the repeated observation of such stereotypes. For both models, the correlation between the frequencies of professions $\mathbf{f}$ and the CV $\mathbf{v}$ is negligible ($|r| < 0.20$). In other words, the model checkpoints do not produce relatively fixed bias scores, for both frequently and infrequently seen professions.

\noindent\textbf{RQ4: Do biases fluctuate for all professions at the same time?}

\begin{figure}[!htbp]
    \centering
    \includegraphics[width=6cm]{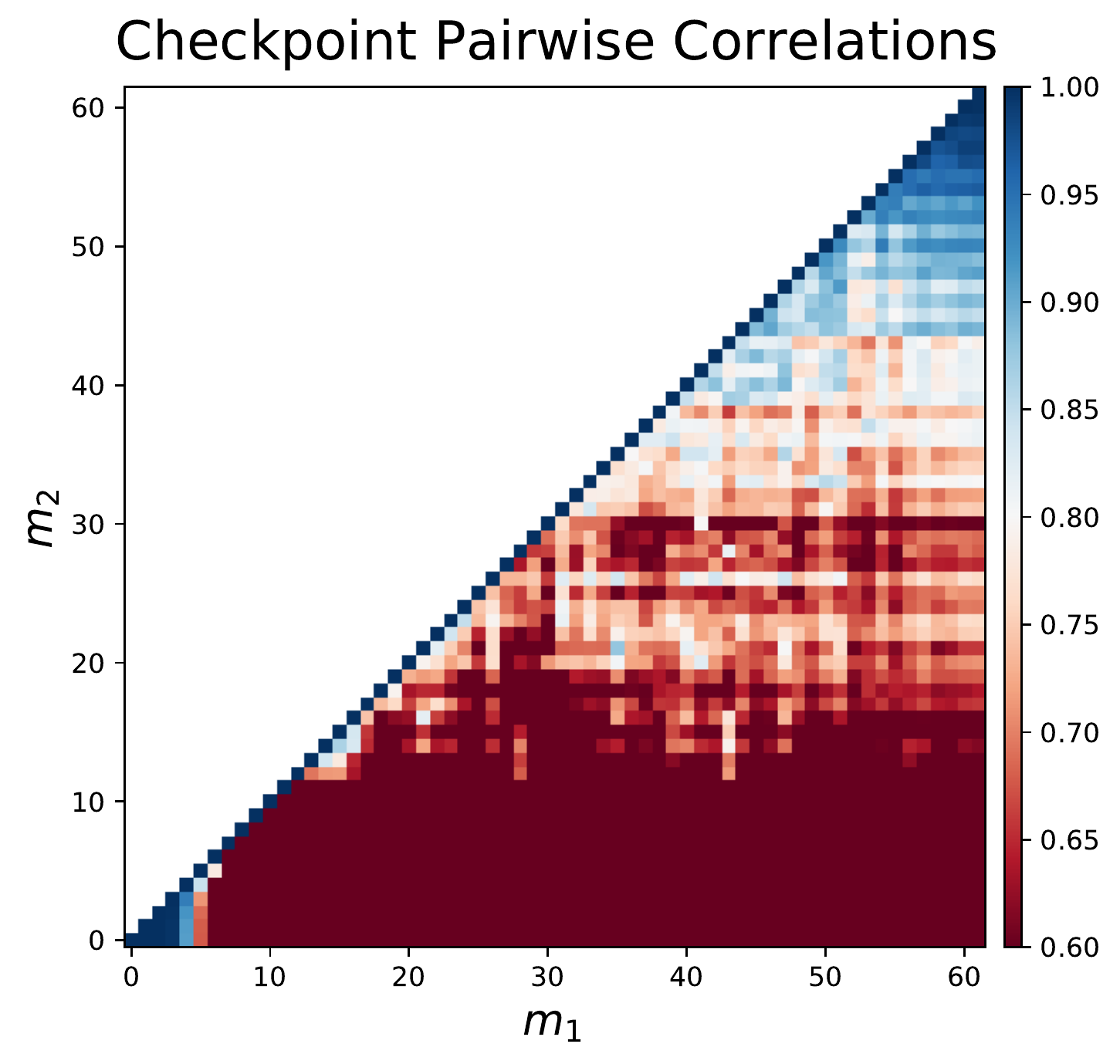}
    \caption{For pairs of RoBERTa checkpoints, the correlations between $\mathbf{N}_{m,:}$ are strong towards the end of pre-training. However, fluctuation in normalized ratios of individual templates cannot be ignored. Correlations below 0.6 have been truncated in this visualization.}
    \label{fig:RQ4}
\end{figure}

Humans link genders with certain qualities such as aggressiveness or caring. As such qualities are shared as the requirements of multiple professions, human stereotypes of professions are reasonably not independent. We ask if models would behave similarly, by calculating the correlations between the pairs of ratio vectors $\mathbf{N}_{m_1,:}$ and $\mathbf{N}_{m_2,:}$ for any pair of checkpoints $m_1$ and $m_2$ in one pre-training run (Figure \ref{fig:RQ4}). In the results, we do observe that correlation is strong ($r>0.80$) between some pairs of steps. However, the high fluctuation of ratios for many professions still weakens correlation after the loss has already plateaued ($m_i \geq k = 36$).

\noindent\textbf{RQ5: How do random seeds influence the fluctuations?}

Human individuals perceive gender stereotypes to various degrees. When calculated from the normalized ratios $\mathbf{N}$, the distribution of the CV apparently shifts when the random seed changes (Figure \ref{fig:RQ12}, bottom row). However, the shift is due to the high variance of prior probabilities estimated from a single template. When calculated from the unnormalized ratios $\mathbf{R}$, this discrepancy between distributions has been largely mitigated (Figure \ref{fig:RQ12}, top row). This result warns against the over-reliance on the template-based prior estimation method.

\begin{figure}[!htbp]
    \centering
    \includegraphics[width=6cm]{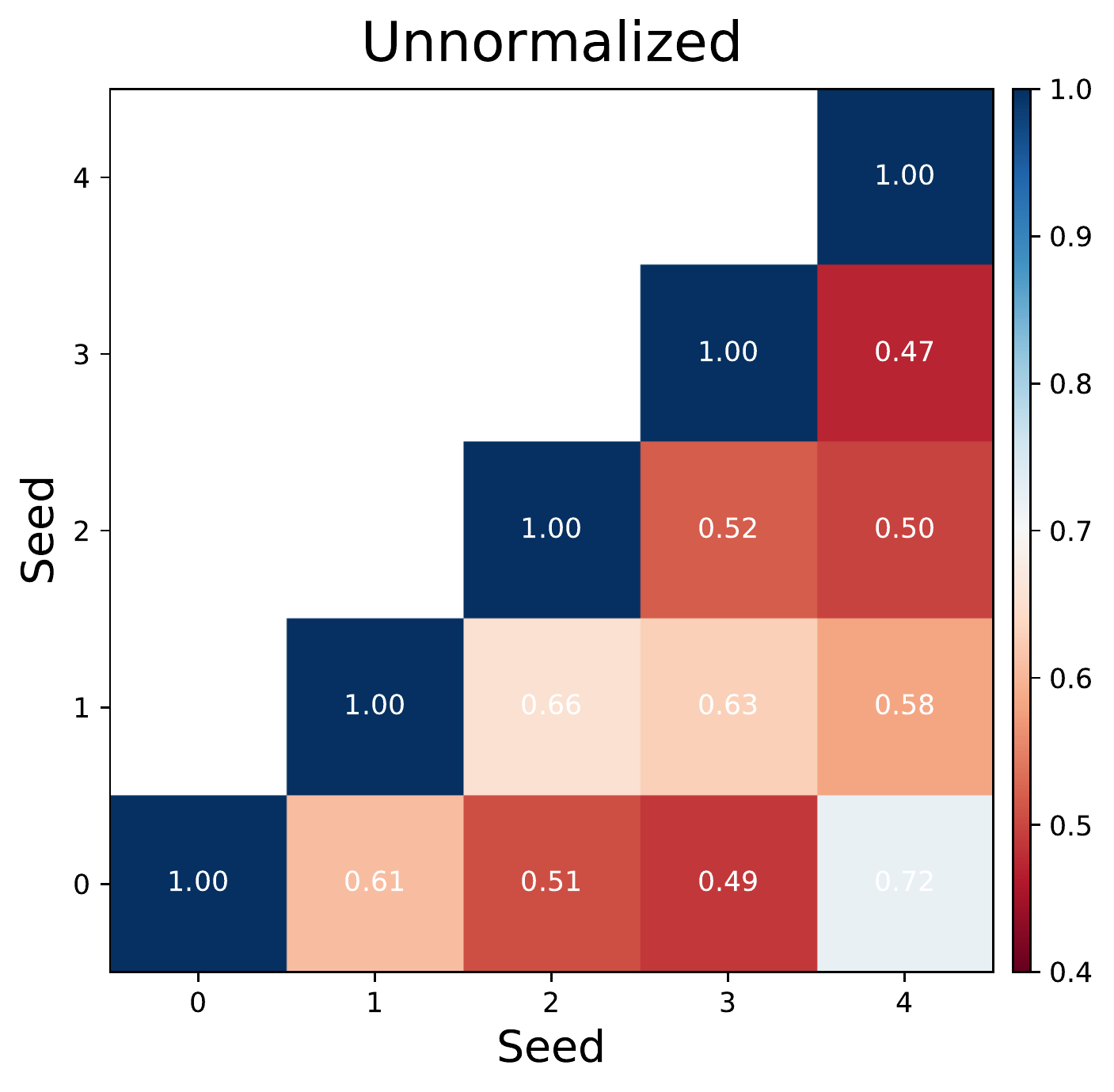}
    \includegraphics[width=6cm]{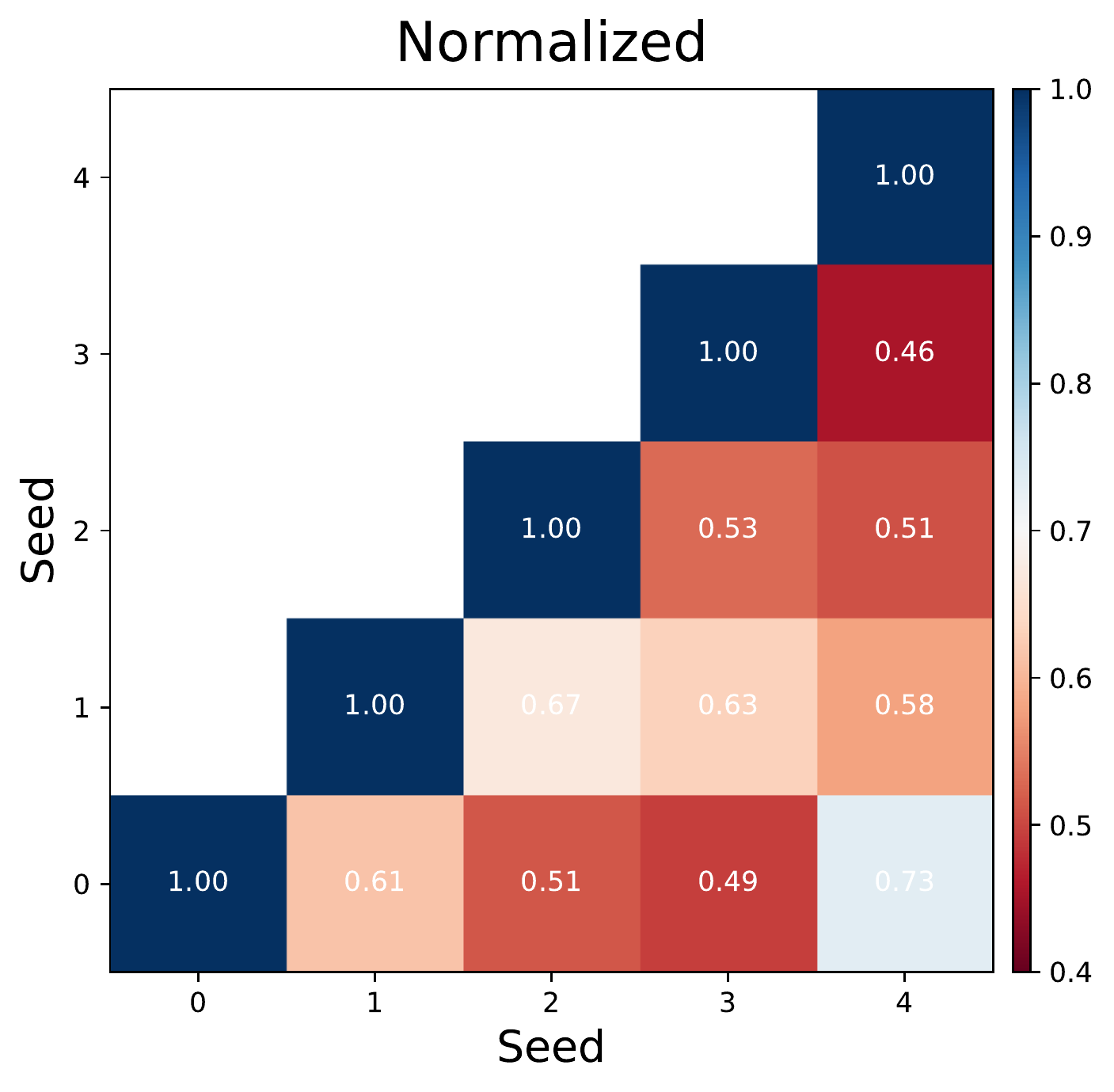}
    \caption{For BERT, different pre-training seeds lead to weakly correlated bias scores for all professions. The results for normalized and unnormalized ratios are different but close. All correlations are above 0.40.}
    \label{fig:RQ5}
\end{figure}

By further computing the pairwise correlation between the averaged ratios $\mathbf{r}$ (or $\mathbf{n}$) of pre-training runs with different random seeds (Figure \ref{fig:RQ5}), we show that models from different runs return considerably different ($r \approx 0.60$) bias scores for individual professions. This result challenges the belief that a single pre-training run would allow a reliable estimation of gender biases.

\section{Limits}

We summarize the major limits of this paper as follows:

\begin{itemize}
    \item There is no direct correlation between the bias scores measured from the templates and the actual bias the model is expected to exhibit in a downstream task \cite{kaneko-etal-2022-debiasing}. Moreover, we only limit our experiment to the one template-based approach of measuring bias on pre-trained MLMs.
    \item We have not discussed auto-regressive models such as ones from the GPT family. The only publicly available pre-training checkpoints are for GPT-2 \cite{Mistral}, which are not expected to behave similarly as the much larger and more popular GPT-3.
    \item We have only discussed the pre-training process on the largest corpora available. On one hand, continued pre-training could use smaller, domain-specific corpora \cite{gururangan-etal-2020-dont}. On the other hand, MLMs can be trained on different pre-training objectives \cite{alajrami-aletras-2022-pre}. Both variations should influence the pre-training dynamics. 
\end{itemize}

These factors limit the generalizability of our conclusions.

\section{Conclusion}

In this paper, we show that when measured by a popular template-based probe, the gender biases in MLMs fluctuate with respect to different random seeds and the number of pre-training steps after the loss has plateaued. Moreover, we provide how these fluctuations can be interpreted in ways that do not necessarily align with intuitions. Specifically, such fluctuations should be taken into account when MLMs are used for capturing biases from a certain corpus, or when biases of MLMs are compared for the evaluation of de-biasing methods.

\bibliography{anthology,custom}
\bibliographystyle{acl_natbib}

\appendix

\section{Data Format}
\label{sec:format}

The data we release are in the following format. Columns in the data frame represent:
\begin{itemize}
    \item The pronoun. The value is either ``he'' or ``she'' for \texttt{bert-base-uncased} and either ``He'' or ``She'' for \texttt{roberta-base}.
    \item The score. The value is between 0 and 1.
    \item The profession. The value is either a string from the list of 923 professions (e.g., ``nurse'') or the mask token used for prior estimation.
    \item The template. The value is a masked template without a pronoun.
    \item The full sentence. The value is a full sentence with a pronoun.
    \item The model name. The value is either \texttt{roberta-base} or \texttt{bert-base-uncased}.
    \item The index of the pre-training seed. The value is in $\{-1, 0, 1, 2, 3, 4\}$ for \texttt{bert-base-uncased} and in $\{-1, 0\}$ for \texttt{roberta-base}. Though not investigated in this paper, the data computed using the public checkpoints from Hugging Face are included and are indexed by $-1$.
    \item The checkpoint. The value is either the number of pre-training steps of a checkpoint or \texttt{NaN}, the latter representing the single available public checkpoint.
\end{itemize}

\section{Starting Point of the Plateau}
\label{sec:plateau}

\begin{figure*}[!htbp]
    \centering
    \includegraphics[width=5cm]{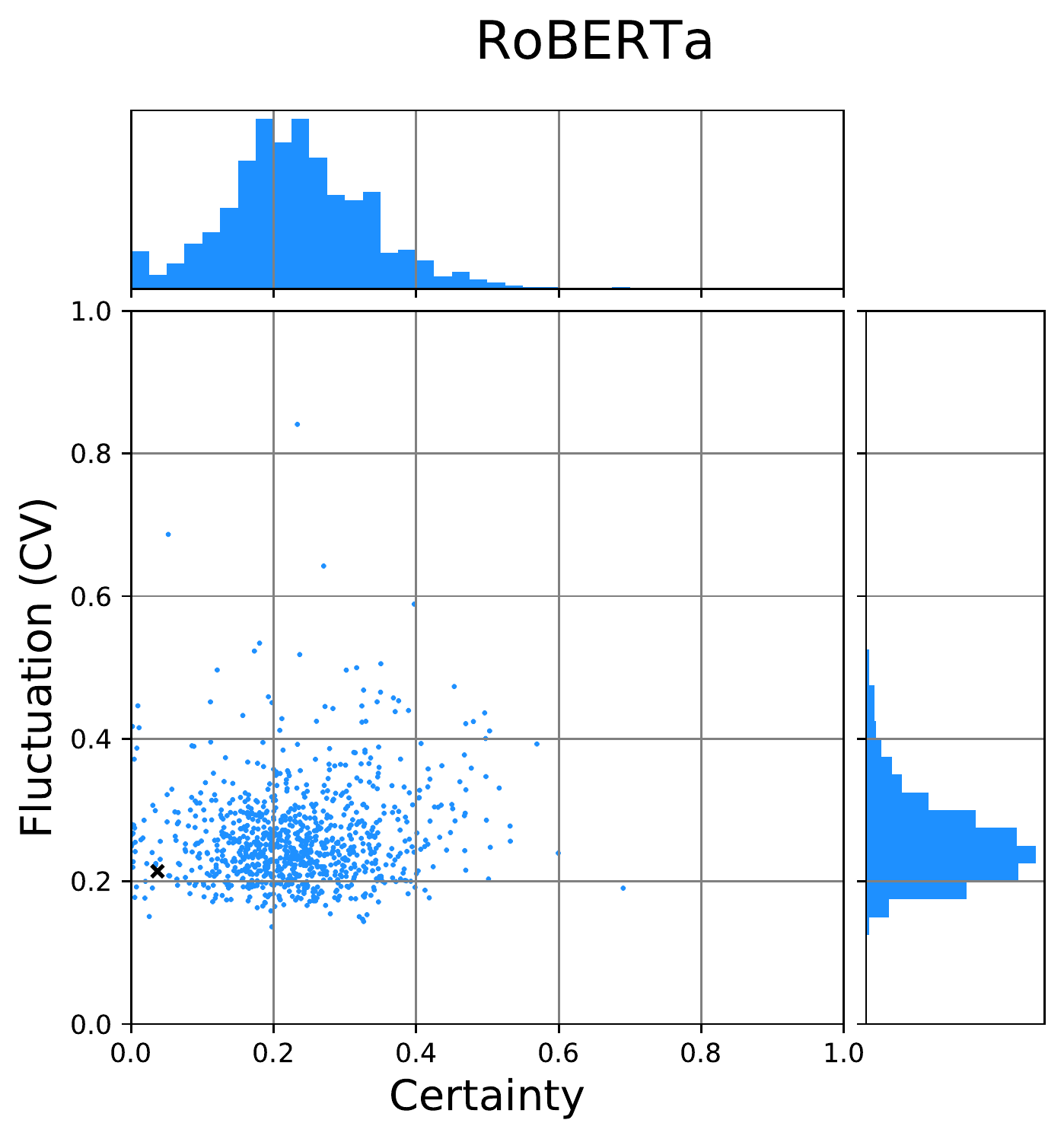}
    \includegraphics[width=5cm]{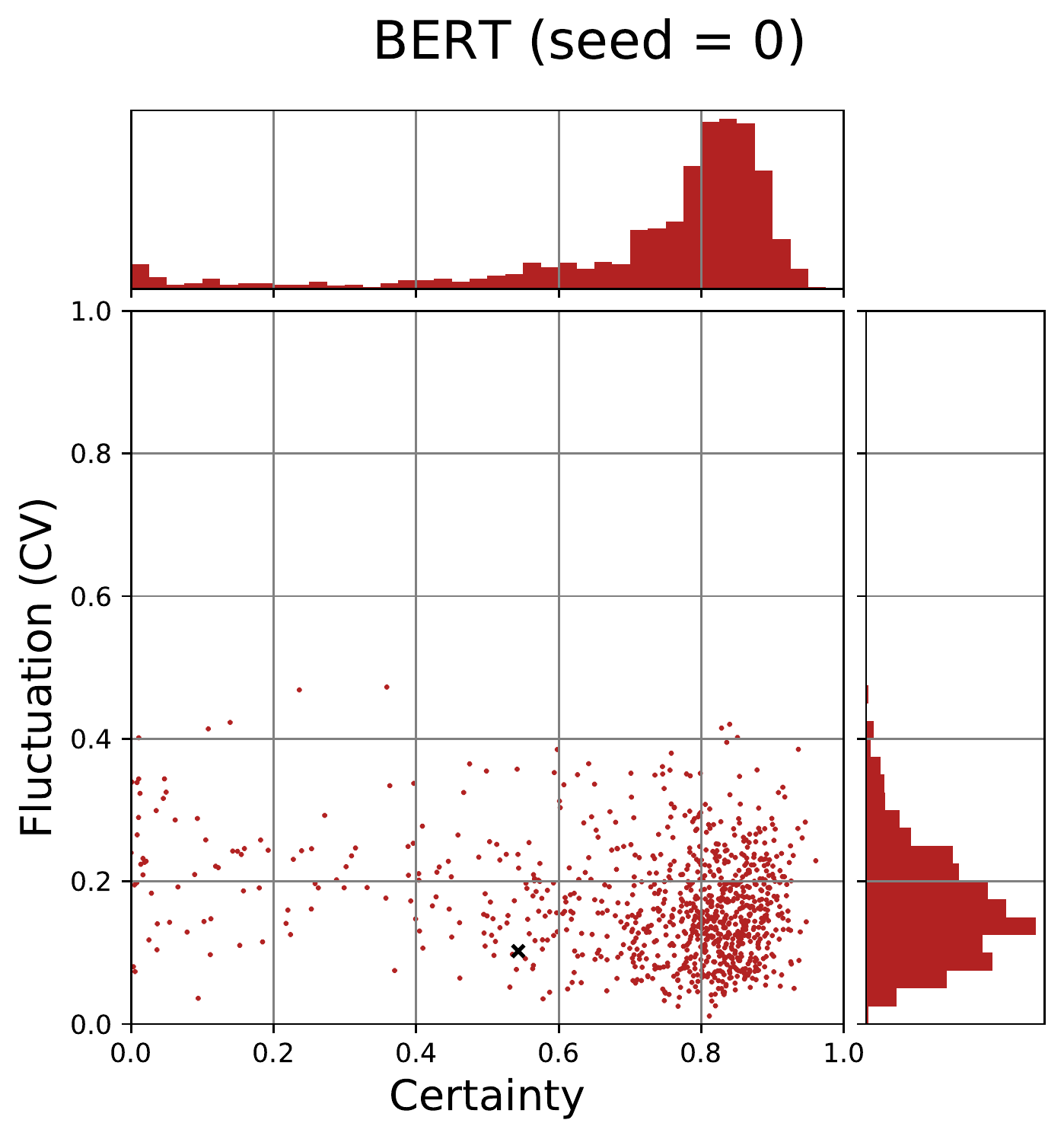}
    \includegraphics[width=5cm]{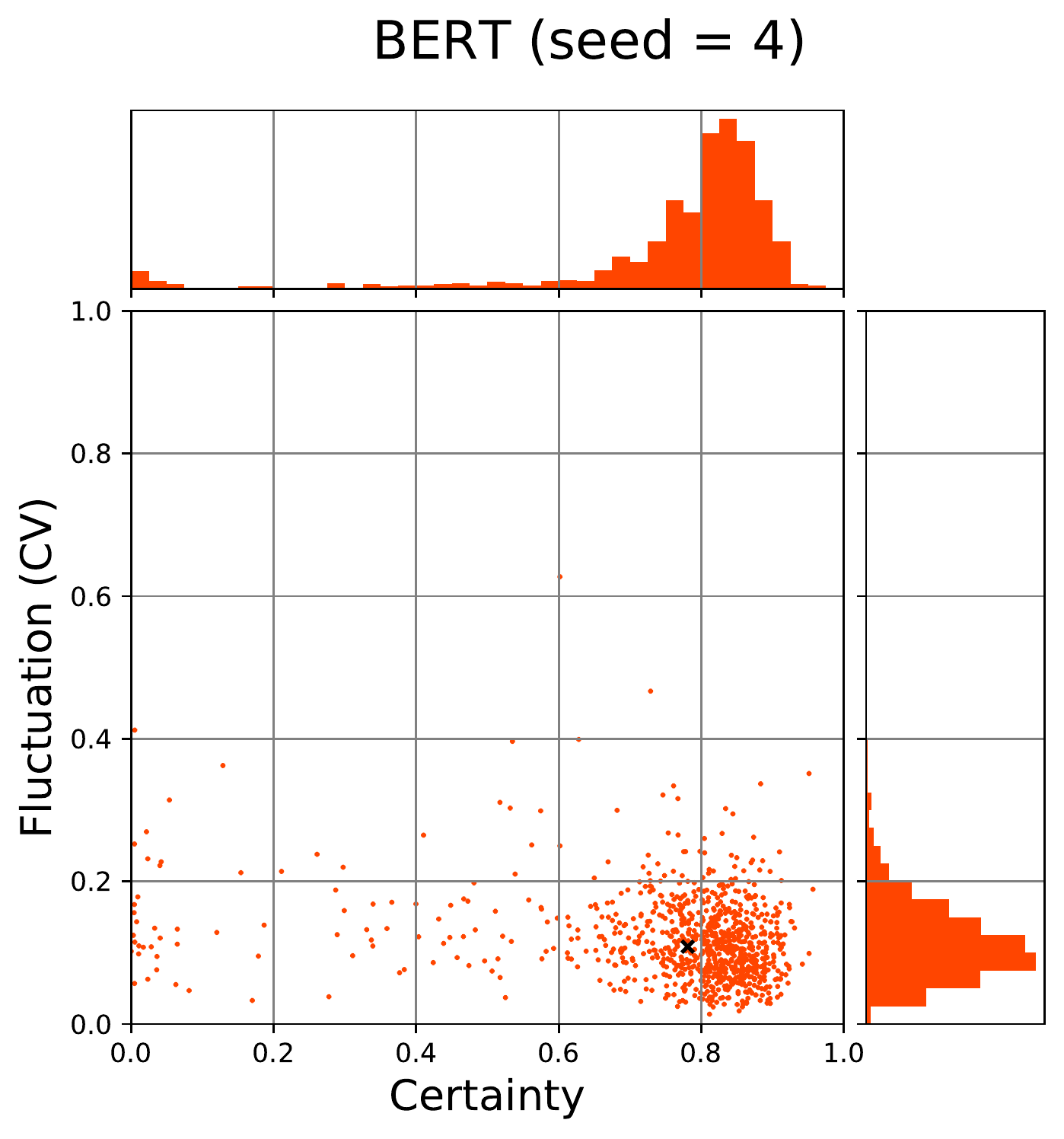}

    \includegraphics[width=5cm]{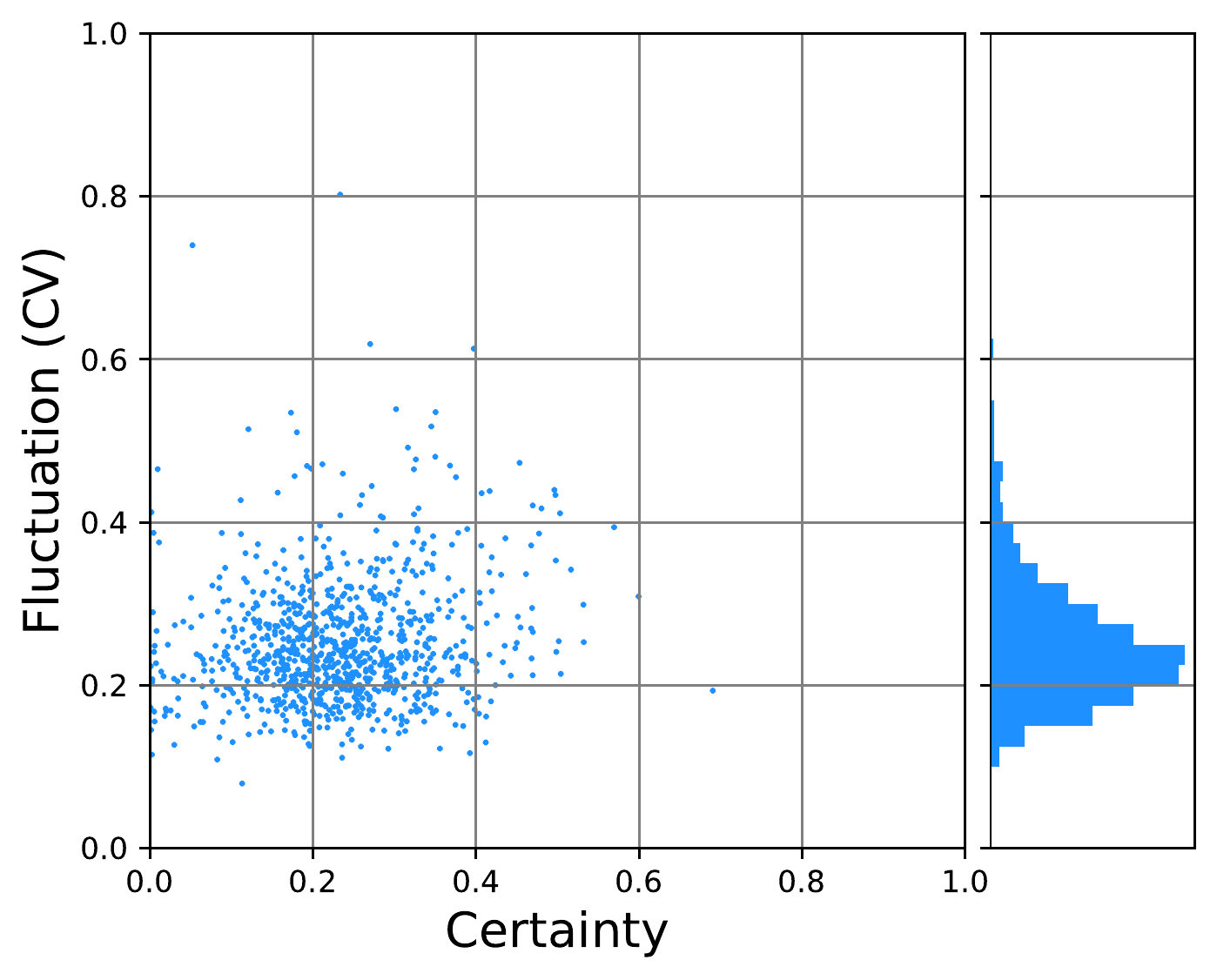}
    \includegraphics[width=5cm]{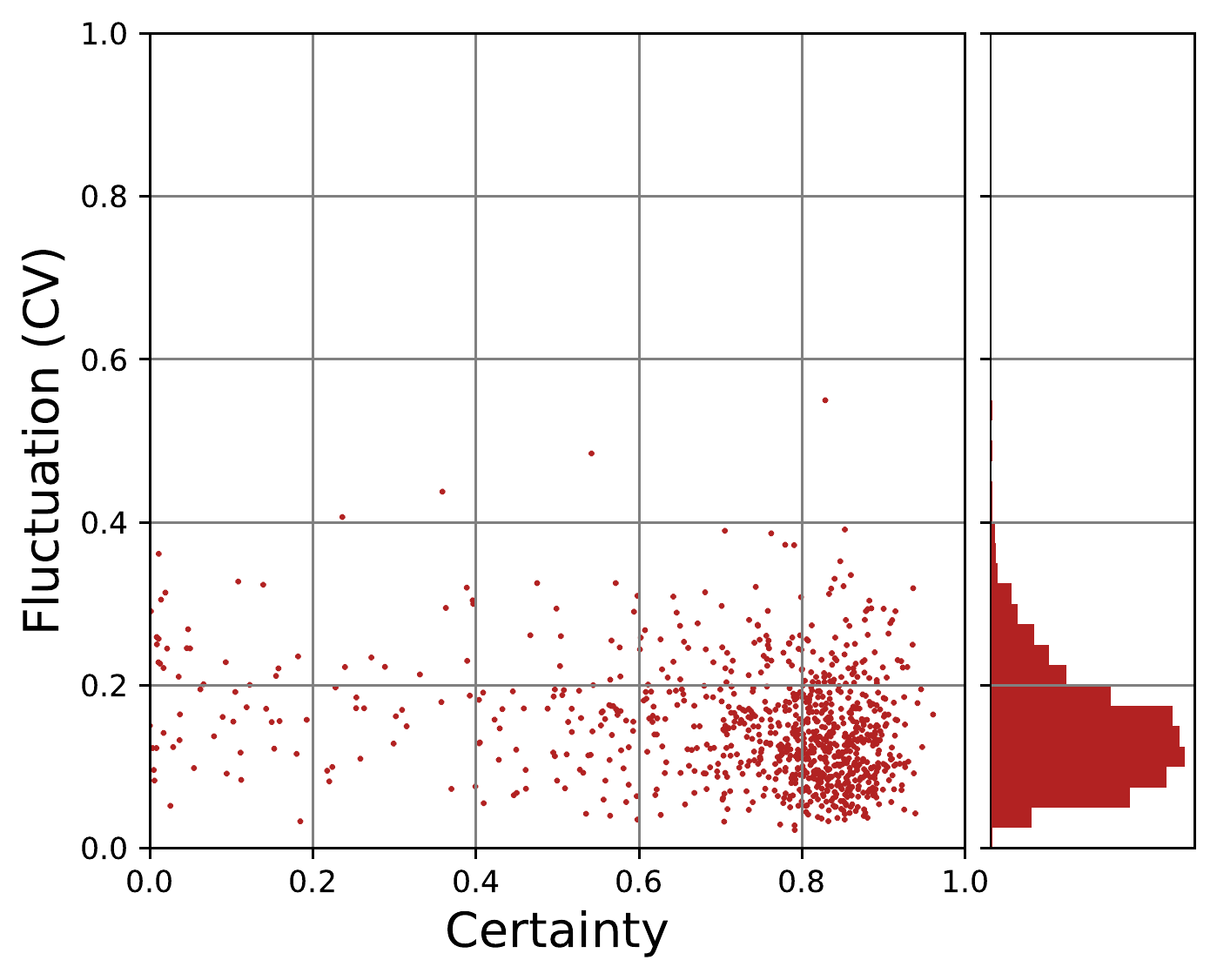}
    \includegraphics[width=5cm]{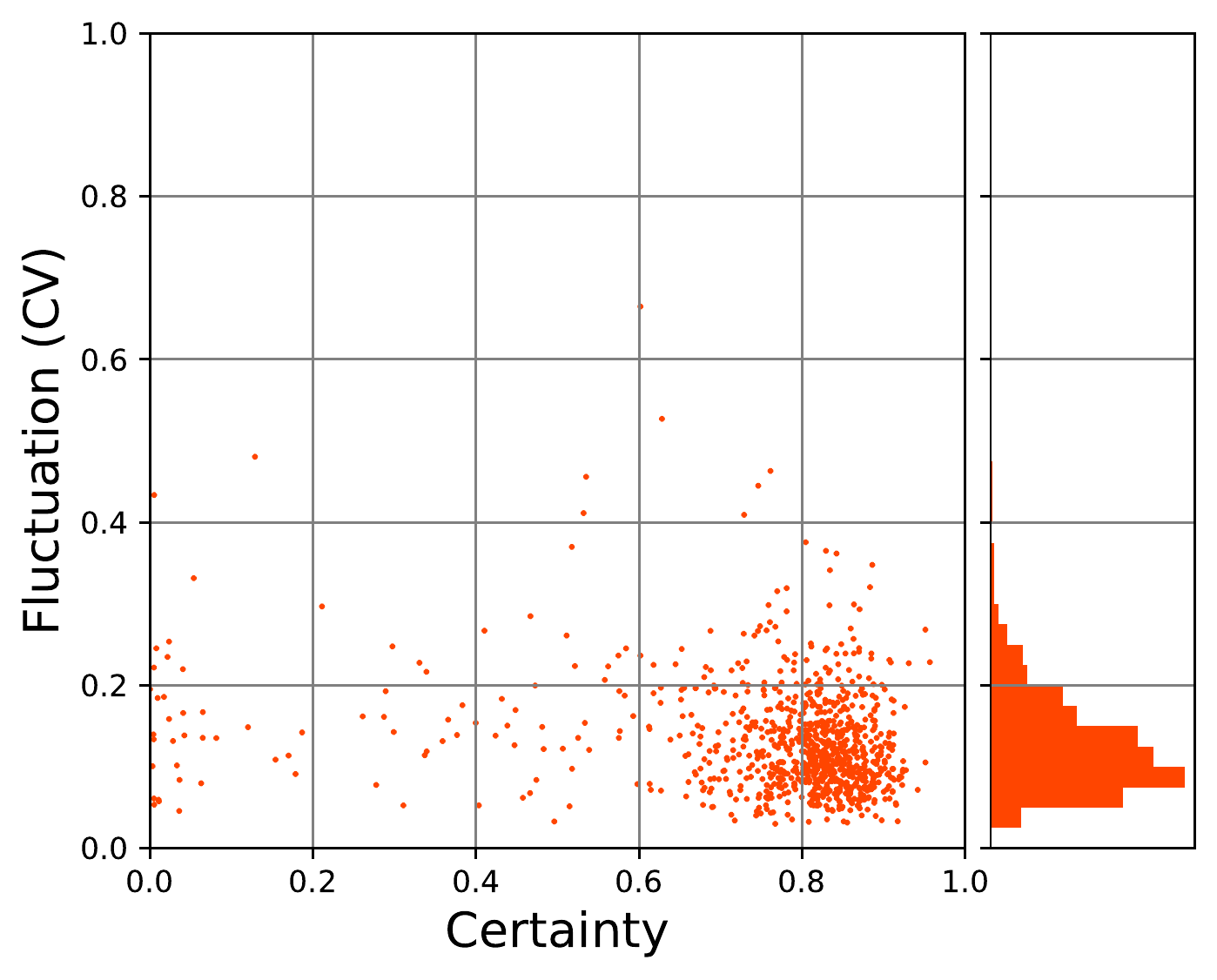}
    
    \caption{The results based on alternative lengths ($k$) of the plateau do not qualitatively differ from the ones based on the reported values (Figure \ref{fig:RQ12}). The top row shows the unnormalized results, and the bottom row shows normalized results.}
    \label{fig:plateau}
\end{figure*}

We additionally provide results (Figure \ref{fig:plateau}) from shorter plateau ranges for both RoBERTa ($k=49$) and BERT ($k=24$). Though smaller than in Figure \ref{fig:RQ12}, the fluctuation still exists and shows no correlation with certainties.

\section{Profession Frequencies}
\label{sec:freqs}

\begin{figure}
    \centering
    \includegraphics[width=5cm]{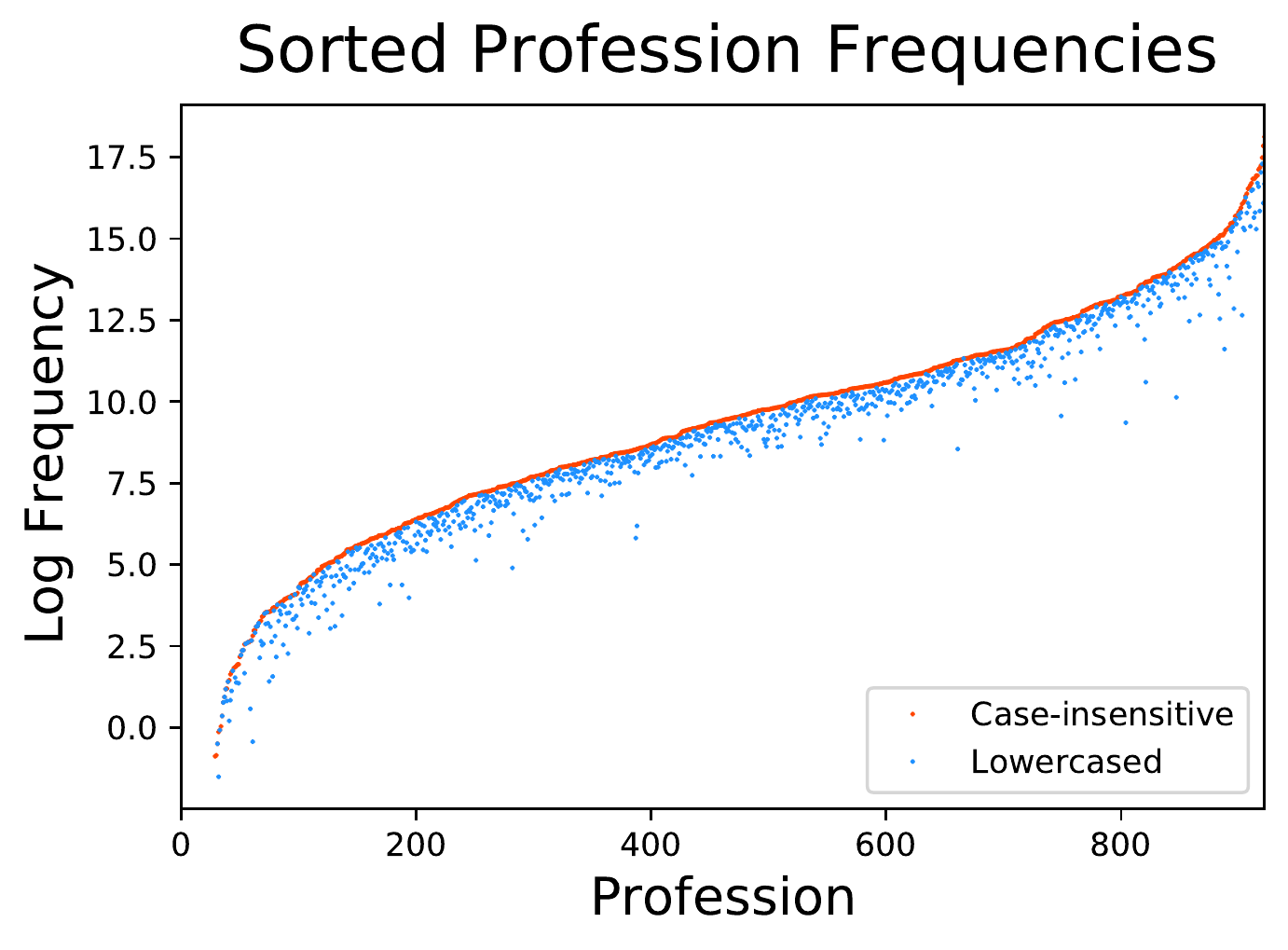}
    \caption{Frequencies of profession names in the BookCorpus. All professions are sorted according to case-insensitive frequencies. }
    \label{fig:profession}
\end{figure}

\begin{figure*}[!htbp]
    \centering
    \includegraphics[width=15cm]{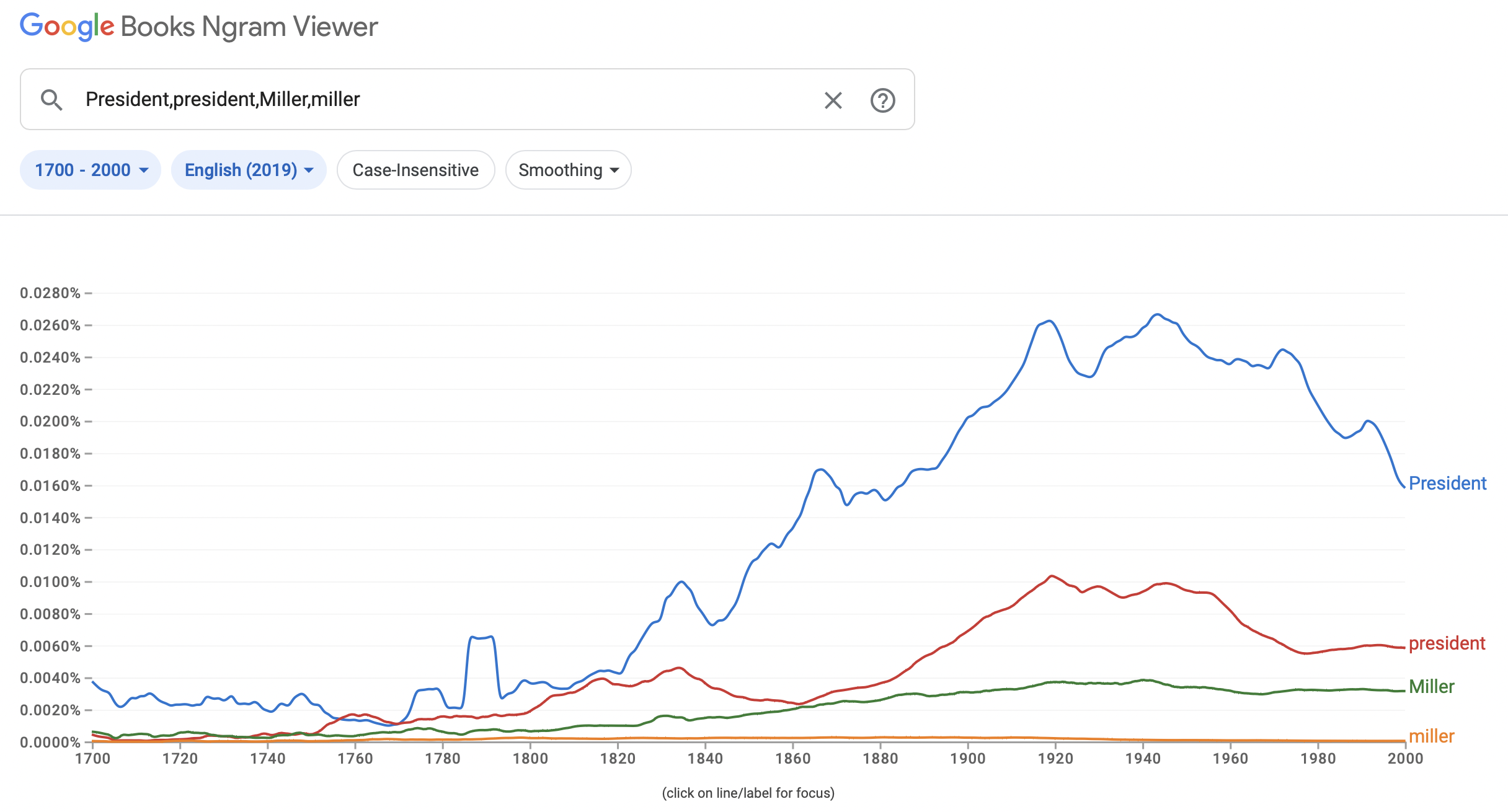}
    \caption{The case of the initial of the profession names influences the relative frequency returned from the Google Ngram API. Uppercase occurrences of certain professions outnumber the lowercased ones for professions such as ``president'' and ``miller''. The screenshot is taken on November 25, 2022.}
    \label{fig:ngram}
\end{figure*}

The sorted profession frequencies are shown in Figure \ref{fig:profession}. The professions with highest frequencies are summarized in Table \ref{tab:profession}. Note that case-sensitivity can lead to very different results for certain professions, such as ``President'' as a title or ``Miller'' as a name (Figure \ref{fig:ngram}).

\section{The Alternative Template}
\label{sec:works-as}

\begin{figure*}[!htbp]
    \centering
    \includegraphics[width=5cm]{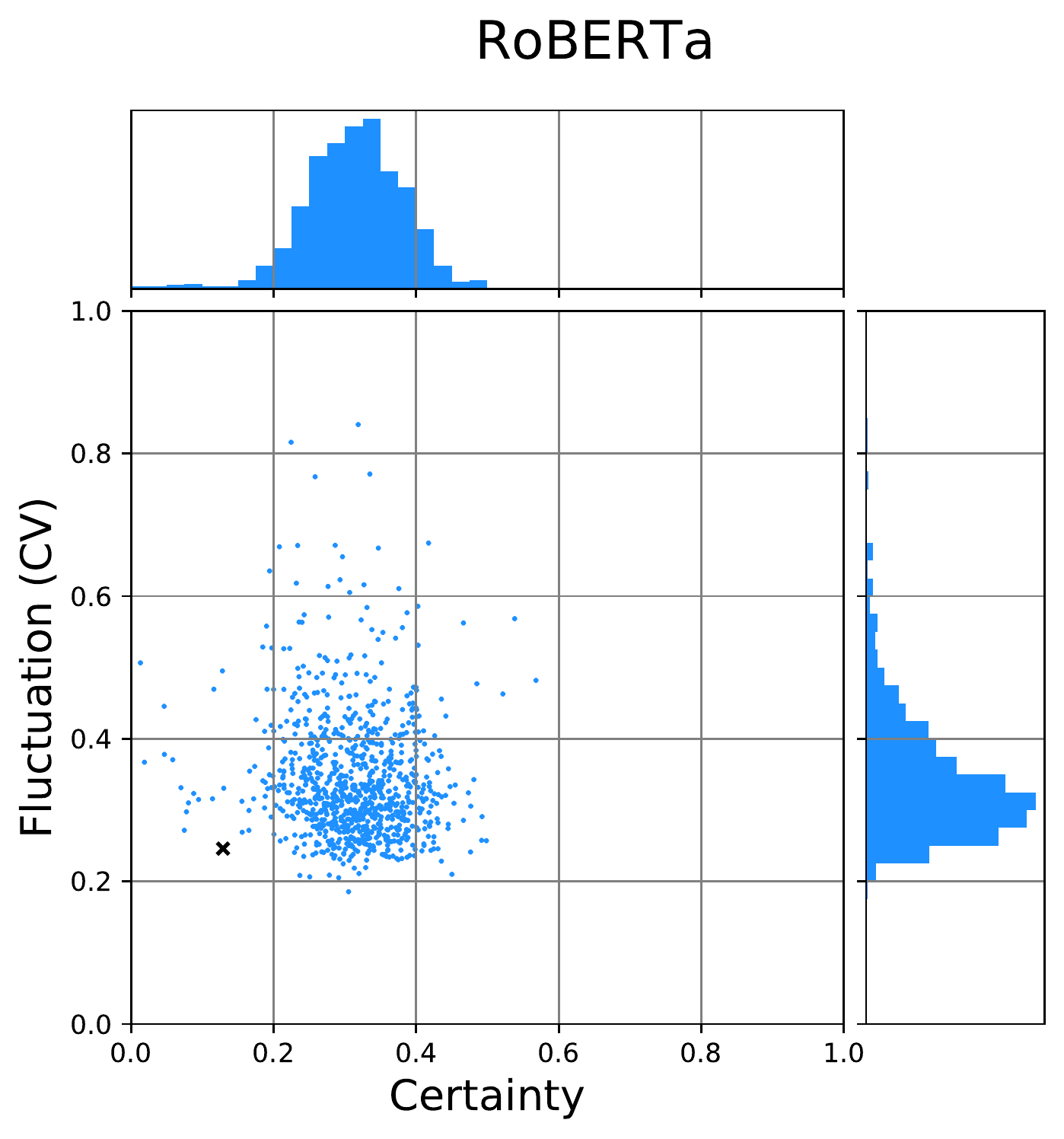}
    \includegraphics[width=5cm]{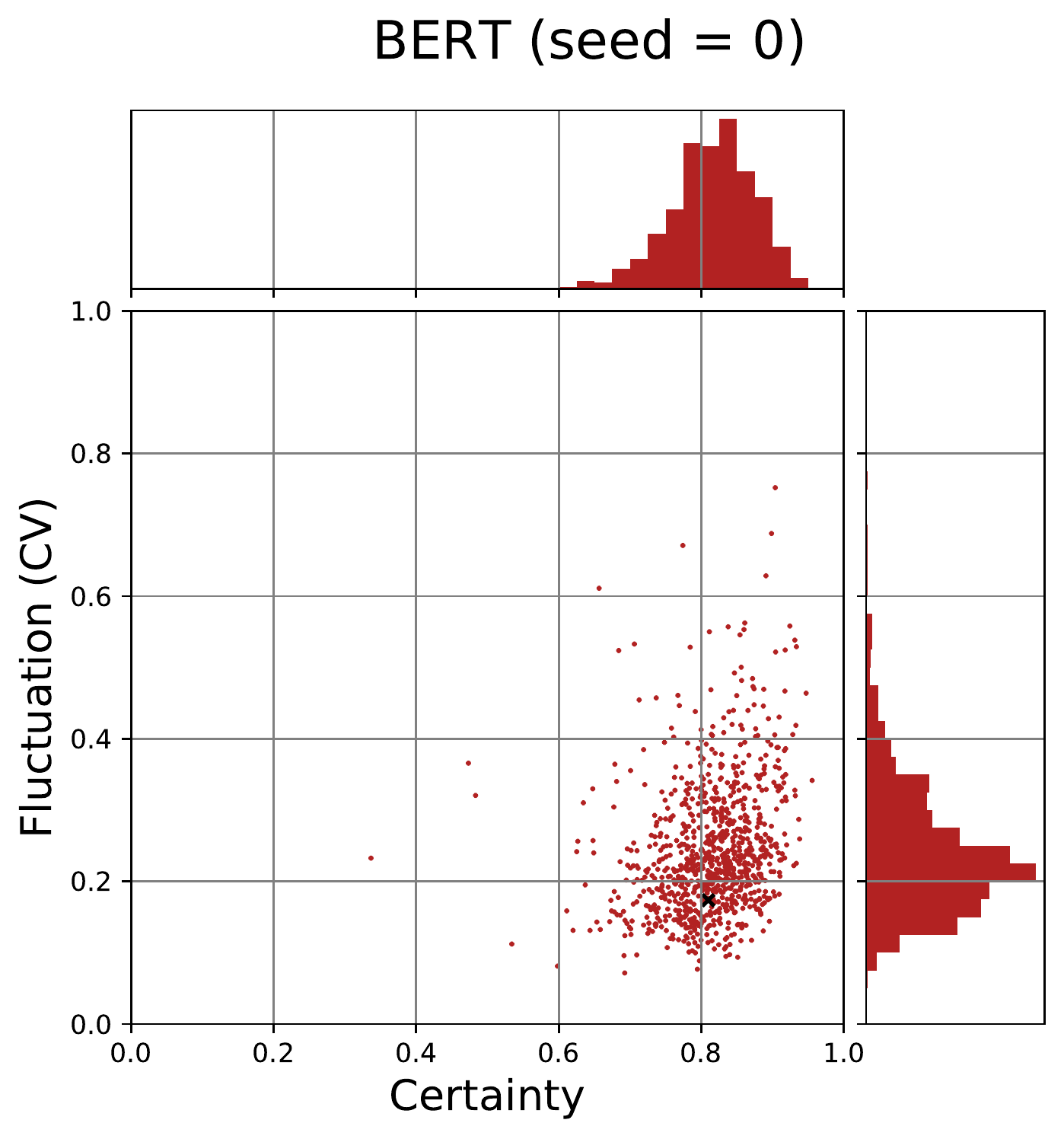}
    \includegraphics[width=5cm]{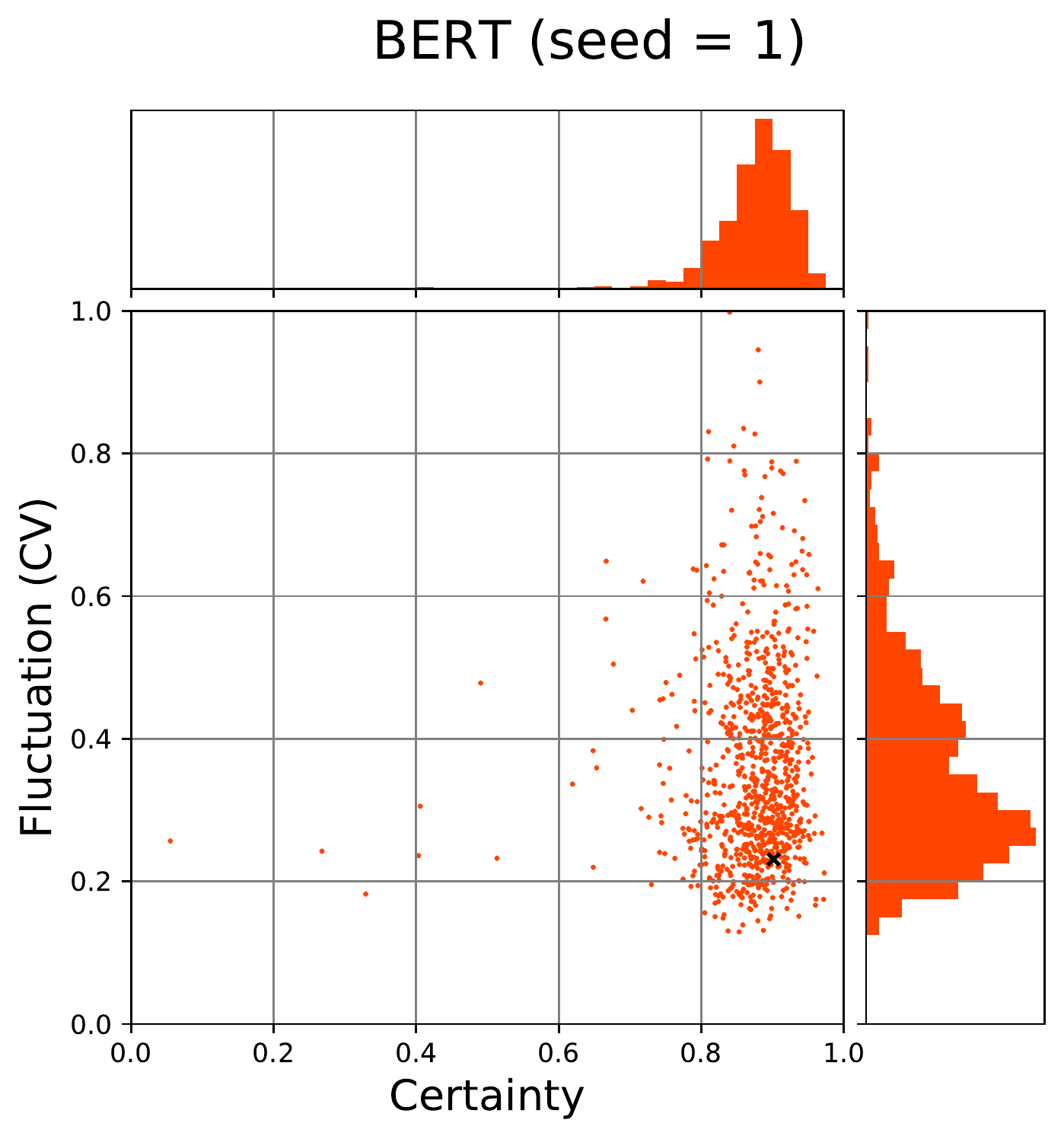}
    
    \includegraphics[width=5cm]{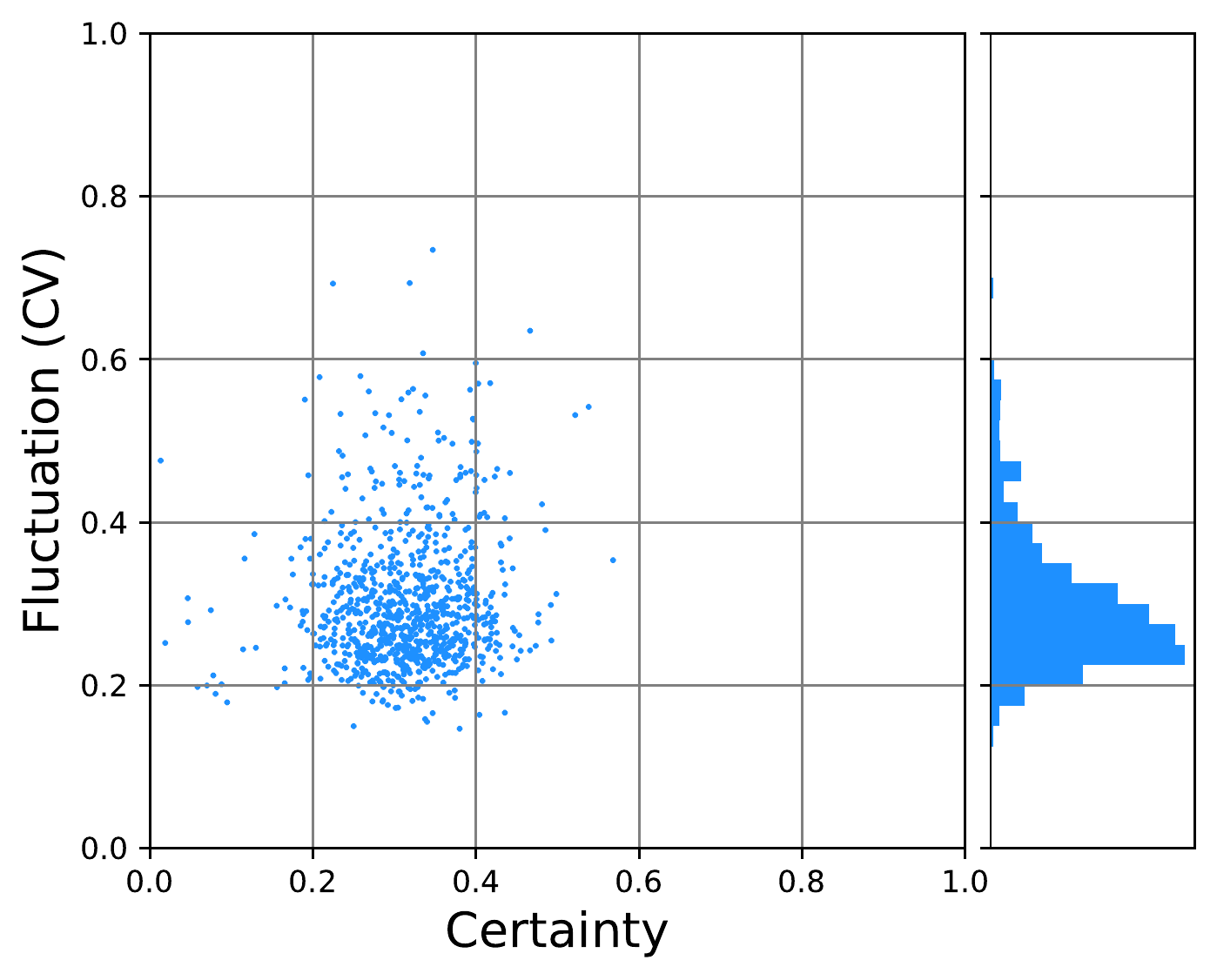}
    \includegraphics[width=5cm]{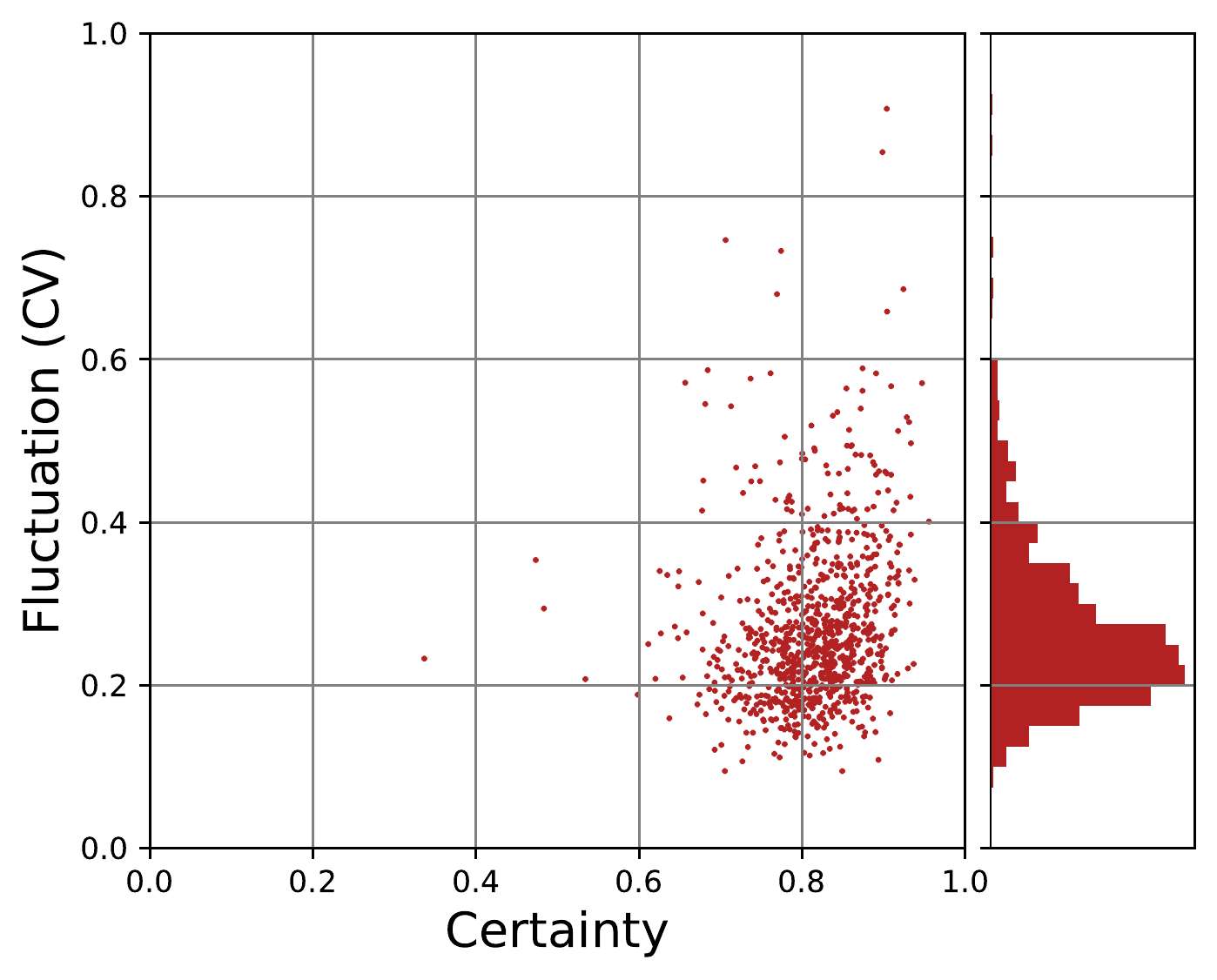}
    \includegraphics[width=5cm]{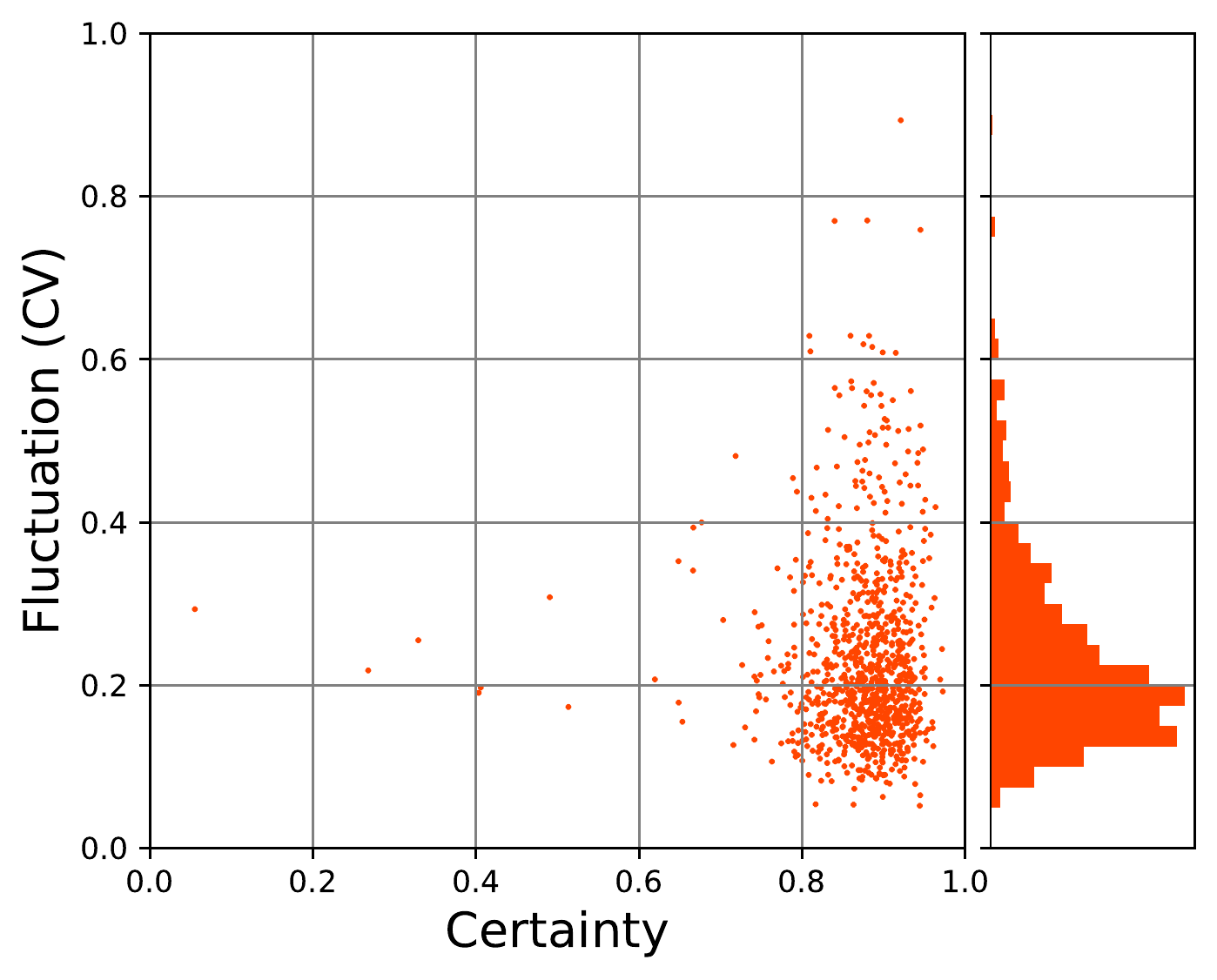}

    \includegraphics[width=5cm]{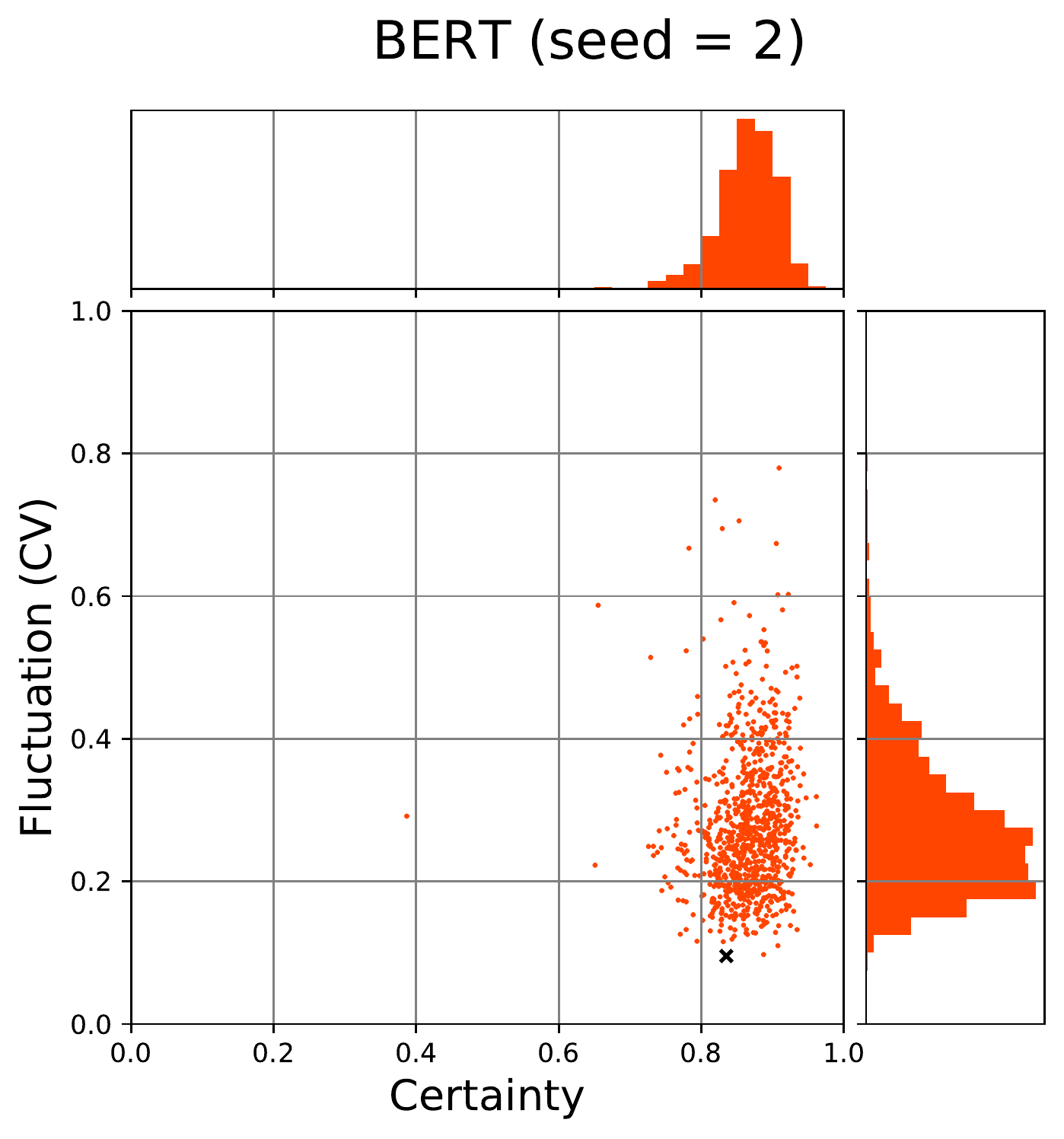}
    \includegraphics[width=5cm]{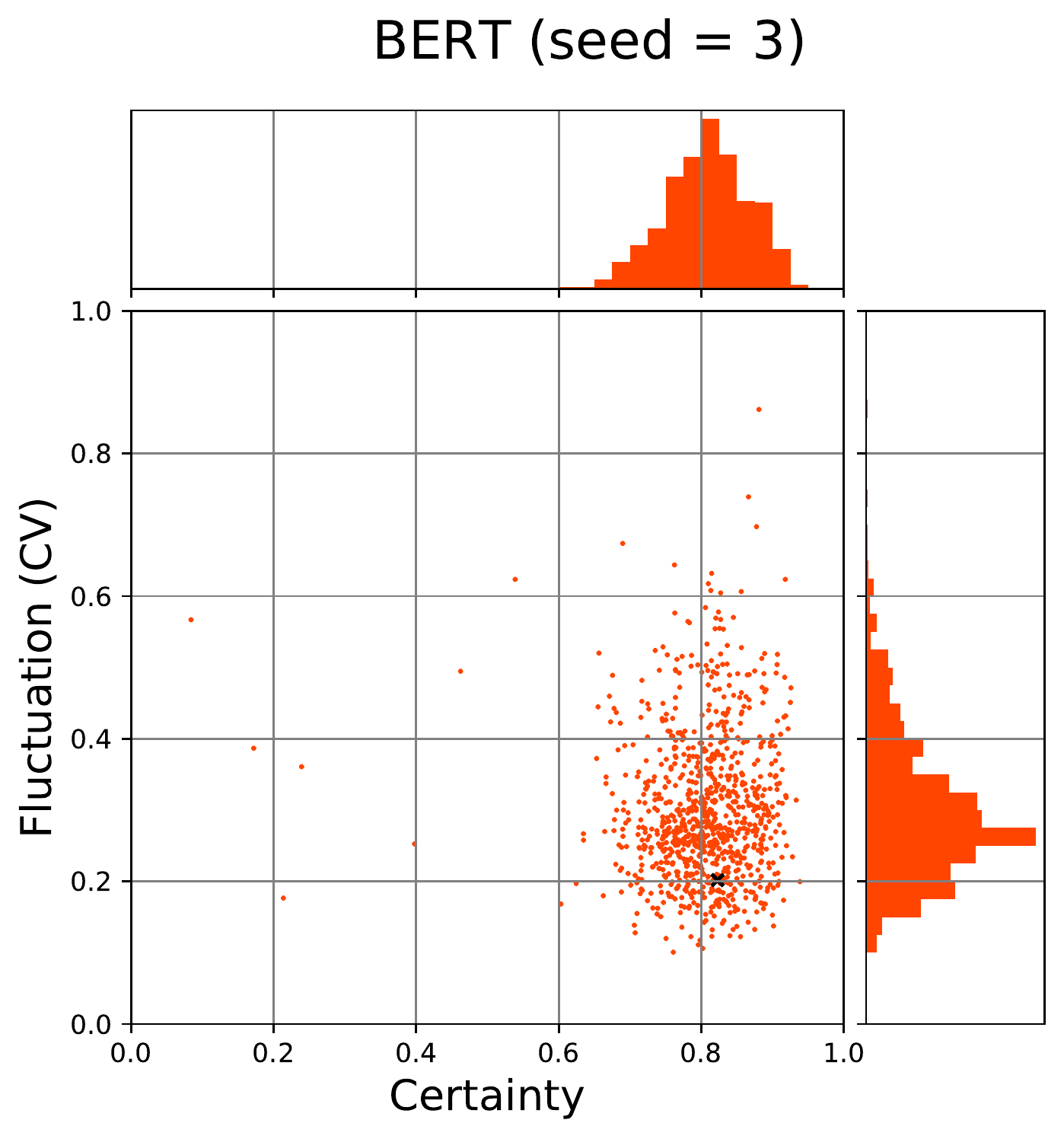}
    \includegraphics[width=5cm]{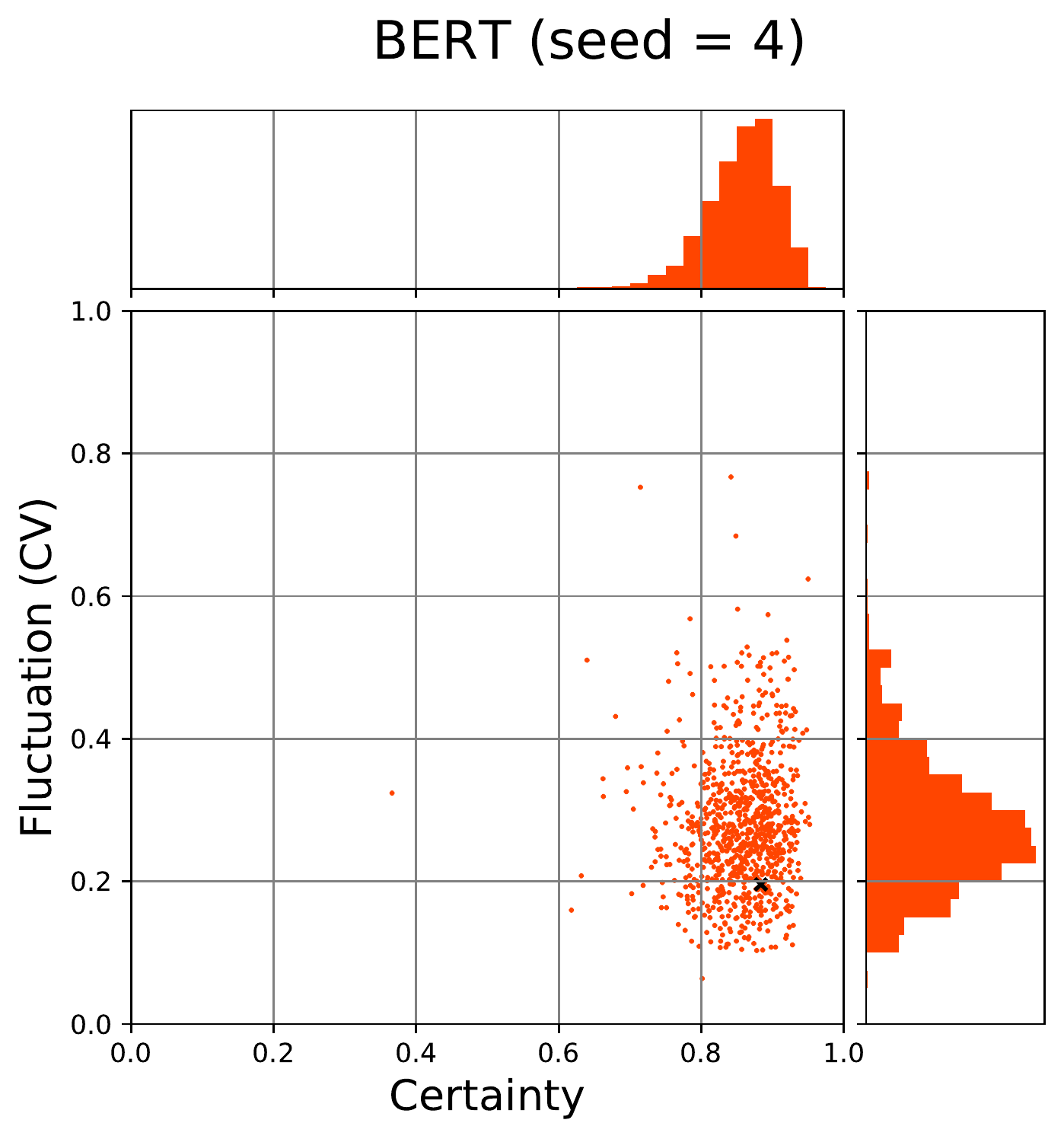}
    
    \includegraphics[width=5cm]{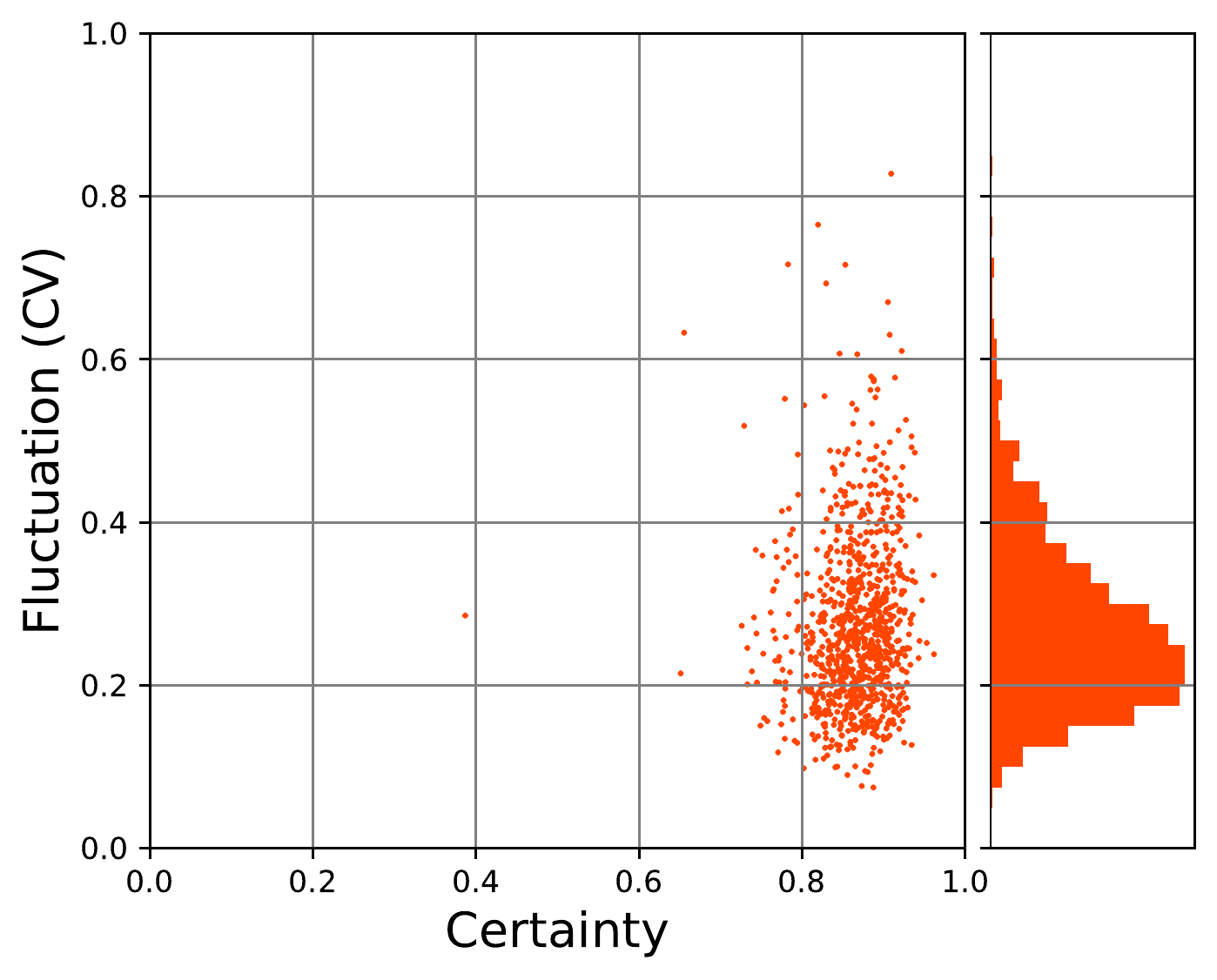}
    \includegraphics[width=5cm]{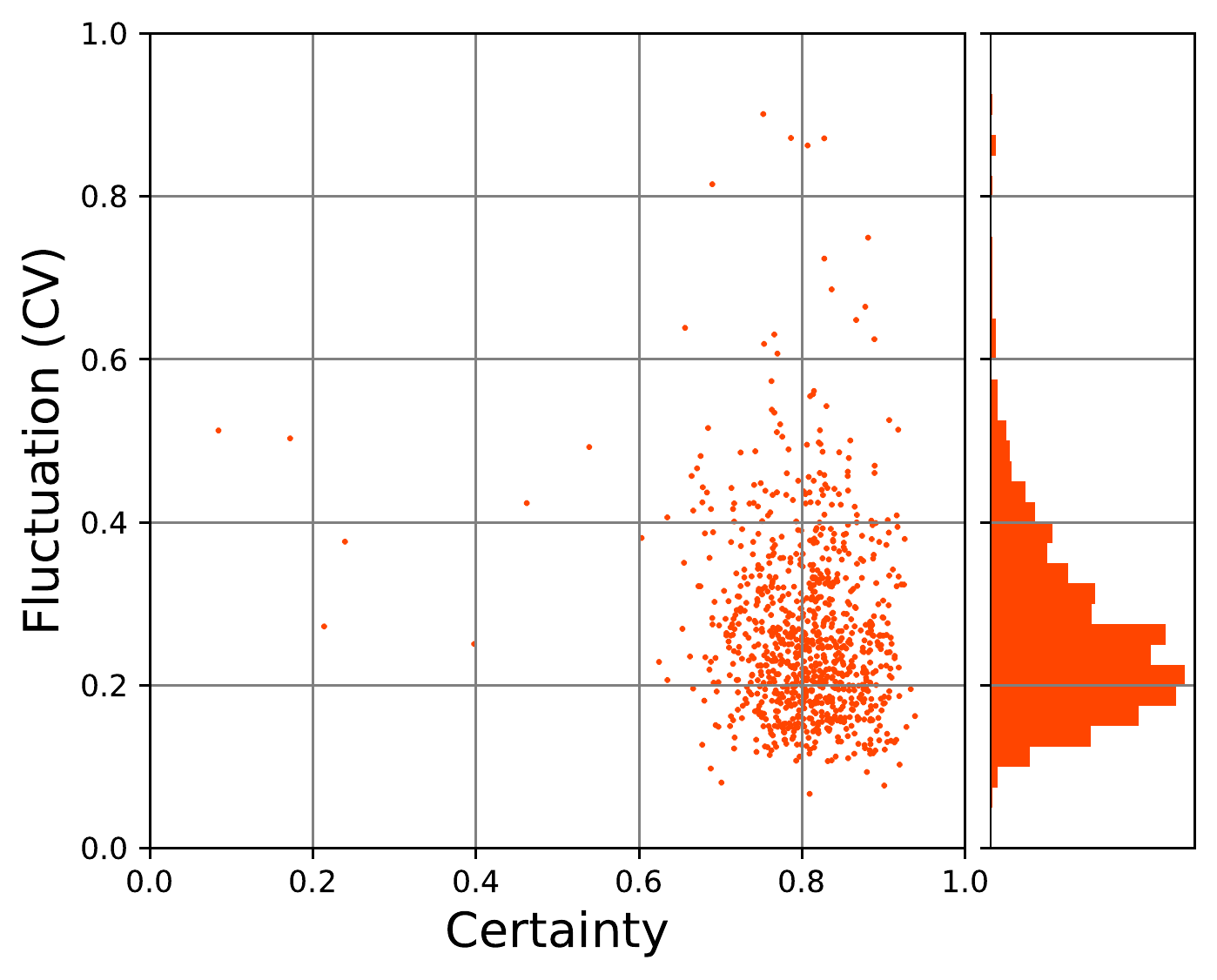}
    \includegraphics[width=5cm]{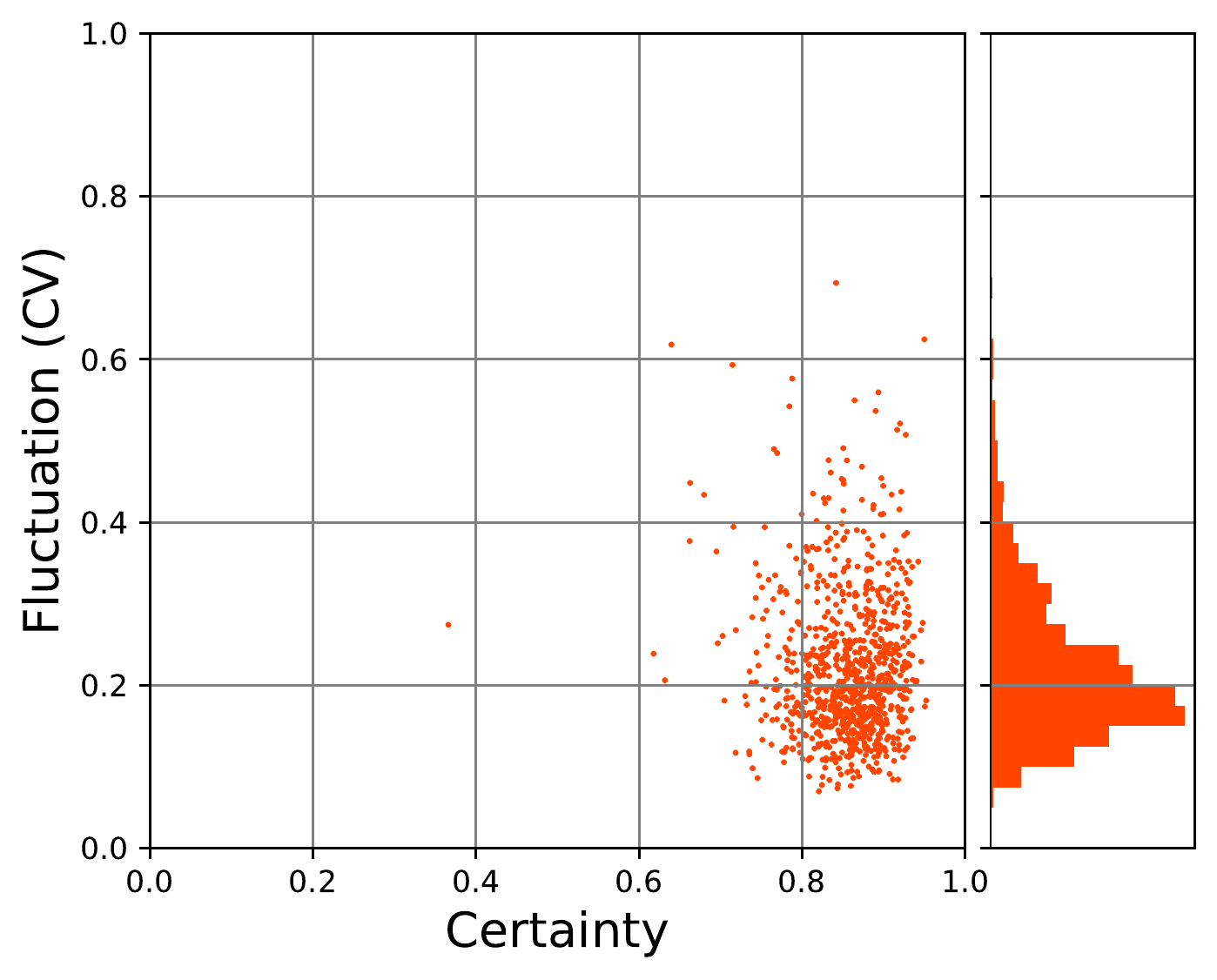}
    
    \caption{The results with the alternative ``works as'' template do not qualitatively differ from the ones with the ``is'' template (Figure \ref{fig:RQ12}). Here results from all 5 random seeds of BERT have been included. The odd rows are unnormalized results and the even rows are normalized results.}
    \label{fig:alt-template}
\end{figure*}

The alternative template with ``works as'' as the \texttt{<VERB>} does not result in qualitatively different results. While we use the scatter plot of fluctuation against certainties as an example here (Figure \ref{fig:alt-template}), all other plots have been released together with the code.

\begin{table}
    \centering
    \begin{tabular}{ll}
        \hline
        Lowercased & Case-insensitive \\
        \hline 
        model & \textbf{president} \\
        author & secretary \\
        official & model \\
        \textbf{president} & author \\
        judge & minister \\
        police & judge \\
        teacher & official \\
        writer & professor \\
        secretary & assistant \\
        guide & governor \\
        clerk & police \\
        minister & teacher \\
        physician & commissioner \\
        assistant & clerk \\
        engineer & guide \\
        host & engineer \\
        governor & writer \\
        farmer & treasurer \\
        artist & superintendent \\
        pilot & \textbf{miller} \\
        \hline
    \end{tabular}
    \caption{The 20 most frequent professions in BookCorpus, ranked using either lowercased or case-insensitive frequencies.}
    \label{tab:profession}
\end{table}

\end{document}